\documentclass[11pt]{article}

\usepackage[final]{acl}

\usepackage{times}
\usepackage{latexsym}

\usepackage[T1]{fontenc}

\usepackage[utf8]{inputenc}

\usepackage{microtype}

\usepackage{inconsolata}

\usepackage{graphicx}
\usepackage{amsmath}
\usepackage{booktabs}
\usepackage{multirow}
\usepackage[table]{xcolor}
\usepackage{makecell}
\usepackage{tabularx}
\usepackage{amssymb}

\usepackage{caption}
\definecolor{camera}{rgb}{0.0, 0.0, 0.0}
\title{Why Are Some Emotions Harder for LLMs? Uncovering the Causal Mechanisms of Emotion Inference via Sparse Autoencoders}

\author{
Bangzhao Shu, Arinjay Singh, Mai ElSherief \\
Northeastern University\\
\texttt{\{shu.b, singh.ari, m.elsherif\}@northeastern.edu}
}

\begin{document}
\maketitle
\begin{abstract}
Large language models (LLMs) are increasingly used in emotionally sensitive human-AI applications, where reliable emotion detection is essential. However, their emotion recognition abilities remain uneven: models often perform well on some emotions while consistently struggling with others. Although recent work has explored emotion mechanisms in LLMs, little is known about why models are weaker on some emotions than others from a mechanistic interpretability perspective. In this work, we investigate emotion-specific biases through the causal mechanisms of emotion inference using sparse autoencoders (SAEs). We systematically identify causal sparse emotion features that drive emotion inference and analyze their sparse causal organization within and across emotions. We show that some emotions, such as surprise and fear, rely on highly concentrated feature sets, whereas disgust exhibits a more distributed sparse causal organization: its causal features are generally weaker, frequently co-activate with features for other emotions, and are often overshadowed by causal features for anger. These representational differences provide a mechanistic explanation for why LLMs struggle more with certain emotions. Finally, we conduct two intervention experiments: targeted steering of weaker causal features to mitigate emotion-specific failures, and global optimization of a steering vector over the identified causal features to improve overall emotion recognition performance.
\end{abstract}

\section{Introduction}

Large language models (LLMs) increasingly interact with humans in emotionally sensitive contexts and demonstrate promising capabilities in emotion recognition and generation~\cite{sabour-etal-2024-emobench,chen-etal-2024-emotionqueen}. A growing body of work evaluates the emotional intelligence (EI) of LLMs across tasks such as emotion recognition and emotional support, showing that model performance varies across settings and tasks~\cite{Yongsatianchot-2023,liu-etal-2021-towards,shu2025fluentunfeelingemotionalblind}. However, LLMs do not perform uniformly across emotions~\cite{koufakou-etal-2024-towards}. Prior studies show that models often perform worse on rare emotions such as disgust and surprise~\cite{saeedi-etal-2025-gt}, and may be biased toward predicting certain emotions, such as anger, while struggling with more subtle emotions~\cite{greschner-klinger-2025-fearful,shu2025fluentunfeelingemotionalblind}. Human annotation studies further suggest that some of these difficulties are not unique to models: surprise is challenging even for human annotators, and disgust often co-occurs with anger~\cite{schuff-etal-2017-annotation}. These findings suggest that emotion-specific errors and biases in LLMs may reflect differences in how emotions are internally represented and processed. Understanding these mechanisms is therefore important for explaining model failures and developing mitigation strategies.

Recent studies have begun to examine emotion processing in LLMs from a mechanistic interpretability (MI) perspective. \citet{tak-etal-2025-mechanistic} show that emotion-related signals can be localized to specific layers or regions of the model. \citet{wang2025llmsfeelemotioncircuits} identify emotion circuits and use them to steer emotional generation, while \citet{lee-etal-2025-large} identify emotion-related neurons that contribute to emotion inference. However, existing work primarily shows that emotion-related representations can be localized, ablated, or controlled. It remains unclear why different emotions are represented with different robustness, why certain emotions are systematically confused, and how these representational differences contribute to imbalanced emotion recognition performance.

Recent advances in MI, particularly sparse autoencoders (SAEs), provide a feature-level view of model representations that enables more interpretable and fine-grained analysis of internal computations. Compared with neuron- and attention-head-level analyses, SAE-based analysis offers a more direct and scalable way to identify, interpret, and intervene on behaviorally relevant features~\cite{cunningham2023sparseautoencodershighlyinterpretable,marks2025sparse}. This makes SAEs especially suitable for studying emotion-specific mechanisms behind biased emotion recognition. We therefore ask the following research questions:
\textbf{RQ1:} How are causal sparse emotion features represented in LLMs, and how do these features differ across emotions?
\textbf{RQ2:} How are causal sparse emotion features organized across layers and emotions, and to what extent do they provide necessary or sufficient control over emotion predictions?
\textbf{RQ3:} Can interventions on causal sparse emotion features mitigate emotion-specific failures and improve emotion recognition more broadly?

Our contributions are as follows. First, we identify causal sparse emotion features that drive emotion inference in LLMs. Second, we show that emotions differ in their sparse causal organization: \emph{fear} and \emph{surprise} are represented by more concentrated and stronger feature sets, whereas \emph{sadness} and \emph{disgust} rely on more distributed and weaker features. Third, we show that \emph{disgust} is weakly represented and is often overshadowed by causal features for \emph{anger}. Fourth, we find strong co-activation between \emph{disgust} and \emph{anger} features, as well as between \emph{surprise} and \emph{joy} features, providing a mechanistic explanation for common emotion confusions. Finally, we show that weakly represented emotions can be improved through targeted feature-level interventions, and that globally optimizing steering vectors over the identified causal features further improves overall emotion recognition performance.

\section{Related Work}

\noindent \textbf{Emotion Recognition and Emotional Bias in LLMs.}
Recent work evaluates the emotional intelligence of LLMs across tasks such as emotion recognition, emotional support, and affective reasoning~\cite{sabour-etal-2024-emobench,chen-etal-2024-emotionqueen,Yongsatianchot-2023,liu-etal-2021-towards,shu2025fluentunfeelingemotionalblind}. Prior studies consistently find that emotion recognition performance varies across emotion categories and evaluation settings~\cite{koufakou-etal-2024-towards,saeedi-etal-2025-gt}. In particular, LLMs may systematically over-predict salient negative emotions such as anger and fear, while struggling with relatively rare or subtle emotions such as disgust and surprise~\cite{greschner-klinger-2025-fearful}. Human annotation studies further suggest that some of these difficulties are inherent to emotion recognition itself, as surprise is challenging even for human annotators and disgust frequently overlaps with anger~\cite{schuff-etal-2017-annotation}.

Together, these studies suggest that emotion recognition errors may be structured rather than random: models show uneven recognition performance across emotion categories, confuse related emotions, and display systematic prediction skews. Yet why such biases arise internally remains unclear, particularly whether they reflect differences in the organization and causal strength of emotion representations inside LLMs.

\noindent \textbf{Interpretability of Emotion in LLMs.}
While most prior research on emotions in LLMs focuses on model outputs, a smaller body of work has begun to investigate the internal mechanisms underlying emotion processing. \citet{hollinsworth-etal-2024-language} show that sentiment can be represented as a linear direction in hidden activations. Using linear probing, \citet{tak-etal-2025-mechanistic} find that emotion-related processing is primarily localized to mid-layers, with multi-head self-attention units playing a dominant role in shaping emotion-related decisions. Recent probing studies further show that emotion-related information appears in both early and intermediate layers depending on the evaluation setting~\cite{palma-etal-2025-llamas,maheswaran-desarkar-2026-unified}. Moving beyond representation analysis, \citet{lee-etal-2025-large} identify neurons associated with specific emotions and observe partial overlap between neurons correlated with anger and disgust. However, these analyses are primarily correlational. \citet{wang2025llmsfeelemotioncircuits} further take a circuit-level perspective and construct emotion circuits that implement emotional computation through causal intervention, demonstrating controllable emotional generation. However, existing neuron- and attention-head-level analyses rely on highly polysemantic units whose semantic roles are difficult to interpret, and interventions on such units can produce unintended or hard-to-predict side effects~\cite{bricken2023monosemanticity,arora-etal-2018-linear,elhage2022toymodelssuperposition}. In addition, prior work remains limited in explaining why different emotions exhibit different robustness and confusion patterns.

Recent work addresses the problem of polysemanticity by using sparse autoencoders (SAEs; see Appendix~\ref{sec:sparse-autoencoders}) to decompose model activations into sparse, approximately monosemantic features~\cite{bricken2023monosemanticity,cunningham2023sparseautoencodershighlyinterpretable,templeton2024scaling}. Building on SAE-based MI and steering methods~\cite{marks2025sparse,he-etal-2025-sae,soo2025interpretable}, our work studies emotion recognition at the sparse feature level. This enables us to decode sparse, approximately monosemantic emotion-related concepts from model activations, allowing us to better study why emotions differ in robustness and confusion patterns, and to perform more interpretable and targeted interventions.

\section{Experimental Setup}

\noindent \textbf{Models, SAEs, and Prompts}.
We use pre-trained SAEs released by Gemma Scope, Gemma Scope 2, and Llama Scope, which provide publicly available SAEs trained on the residual streams of all transformer layers for Gemma-2, Gemma-3, and Llama-3.1-8B models~\cite{lieberum2024gemmascopeopensparse,he2024llamascopeextractingmillions,mcdougall2025gemmascope2}. We include Gemma-2-2B, Gemma-2-9B, Gemma-3-4B-PT, Gemma-3-4B-IT, Gemma-3-12B-PT, Gemma-3-12B-IT, and Llama-3.1-8B in our study, as these are among the few models with publicly available full-layer pre-trained SAEs and sufficient capability for emotion recognition tasks~\cite{gemmateam2024gemma2improvingopen,grattafiori2024llama3herdmodels,gemmateam2025gemma3technicalreport,deng-etal-2025-unveiling,he-etal-2025-sae}. We use 16K-width SAEs for all Gemma models. For Gemma-3 models, we evaluate both the \texttt{big} and \texttt{small} sparsity settings, to explore whether sparsity affects the concentration and entanglement of emotion-related features. For Llama-3.1-8B, we use the 32K-width SAE. Throughout this work, we focus exclusively on residual-stream SAEs, since the residual stream integrates information from both attention and multi-layer perceptron (MLP) sublayers and therefore provides a natural locus for studying high-level distributed representations such as emotions~\cite{zhang2026locatesteerimprovepractical}.

To extract hidden-state activations for emotion inference while minimizing prompt-induced stylistic and instruction-following effects, we adopt a minimal zero-shot prompt:
\texttt{In this text: \emph{[input text]}, the emotion implied is:}

Following prior work~\cite{zhu-etal-2021-topic,tak-etal-2025-mechanistic}, we frame emotion inference as a classification problem and evaluate model outputs by restricting logits to six emotion labels (anger, joy, sadness, fear, surprise, and disgust), rather than relying on open-vocabulary generation.

\noindent \textbf{Datasets}.
To construct generalizable sparse feature representations for emotion inference, we draw on multiple emotion recognition datasets with self-reported emotion labels, which better preserve ecological validity~\cite{pennebaker1986confronting,frattaroli2006experimental,davitz2013language}. Specifically, we include ISEAR, enISEAR, enVent, and EXPRESS~\cite{scherer1994evidence,troiano-etal-2019-crowdsourcing,troiano2023envent,shu2025fluentunfeelingemotionalblind}. Details of the data preprocessing are provided in Appendix~\ref{sec:appendix-data-prep}. 
Across all datasets, we focus on six basic emotions: anger, sadness, joy, disgust, surprise, and fear, following Ekman’s theory of basic emotions~\cite{ekman1999basic}. This taxonomy is widely used in NLP and affective computing due to its cross-cultural validity and conceptual simplicity~\cite{jm3}. The final combined dataset contains 9{,}870 instances.

\section{RQ1: Identifying Causal Sparse Emotion Features}

We identify causal sparse emotion features in three steps. We first select candidate emotion-related features from SAE activations, then estimate their causal effects through ablation, and finally identify a set of emotion-selective features whose causal effects are concentrated on particular emotions.

\begin{figure*}[t]
    \centering
    \includegraphics[width=\textwidth]{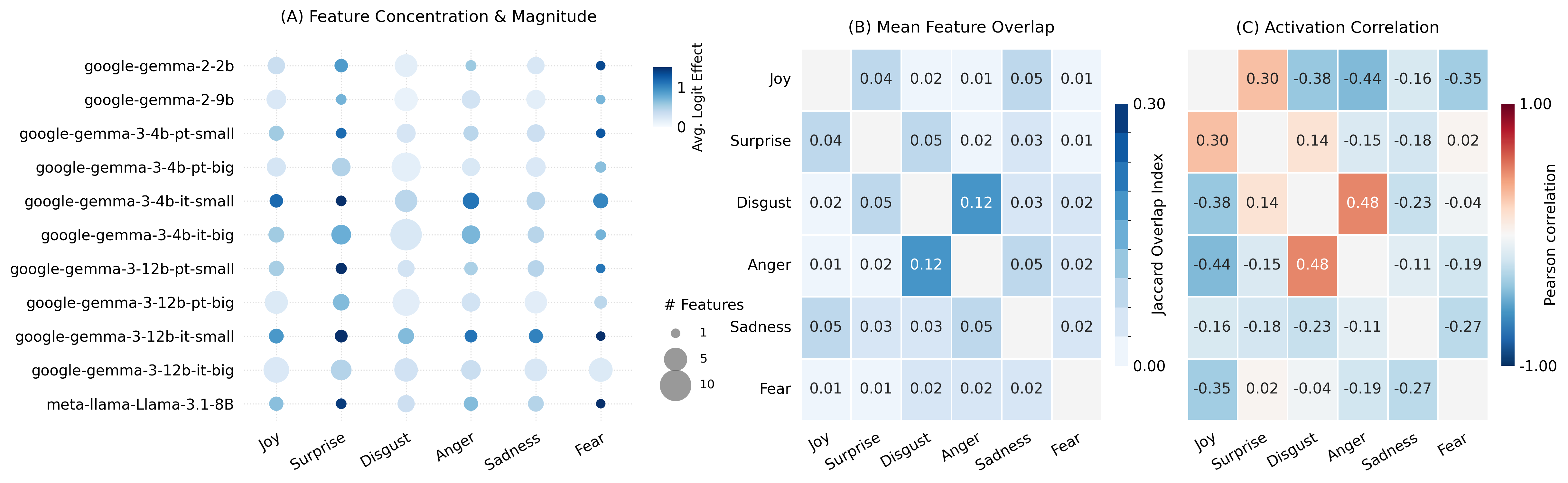}
    \caption{Internal dynamics of causal sparse emotion features across models.
    (A) Feature concentration and causal magnitude by emotion and model, where circle size indicates the average number of selected features per layer and color indicates average causal effect on the target emotion logit.
    \textcolor{camera}{(B) Mean cross-emotion feature overlap, quantified using the Jaccard index (JI) between selected emotion feature sets, with the highest mean JI observed between \emph{anger} and \emph{disgust}.}
    (C) Cross-emotion Pearson correlations of mean activation strength, showing positive correlations for \emph{disgust}--\emph{anger}, \emph{surprise}--\emph{joy}, and \emph{disgust}--\emph{surprise}.}
    \label{fig:figure1}
\end{figure*}

\noindent \textbf{Feature Activation Selection}. We input emotion recognition prompts into the model and extract the hidden states (h) from all layers for each instance at the final token. We focus on the final token because it integrates the preceding context and directly determines the next-token emotion prediction~\cite{oh-schuler-2023-token}. \textcolor{camera}{Recent work on emotion inference also found that emotion information and causal influence increased toward the final input token relative to nearby positions~\cite{tak-etal-2025-mechanistic}.} For each input instance, we collect the final-token hidden states \(h^{(\ell)}\) from all layers \(\ell\) across the dataset. We then encode each hidden state using the corresponding residual-stream SAE to obtain sparse feature activations \(f^{(\ell)}\).

To identify salient and consistently activated emotion-related features while reducing noise, we apply a two-stage filtering process. First, we remove infrequently activated features that activate in less than (1\%) of the dataset. Second, for each emotion \(e\), we compute the mean activation of each feature across all instances labeled with \(e\). We retain the features whose mean activation ranks within the top \(\tau_{\text{act}} = 0.1\) (top 10\%) for that emotion. This procedure selects features that are strongly and consistently associated with a given emotion category. We denote the resulting feature set for emotion \(e\) as \(\mathcal{F}_e\). 


\noindent \textbf{Causal Effect Estimation}. While feature activations reveal correlations with emotions, they do not indicate whether a feature causally contributes to the model’s prediction. We estimate the causal contribution of each feature in \(\mathcal{F}_e\) to the model’s emotion prediction, following prior sparse feature circuits work~\citep{marks2025sparse}. For each feature \(f_i \in \mathcal{F}_e\), we perform feature ablation by zeroing its activation and measuring the change in the output logit corresponding to emotion \(e\), denoted as \(\operatorname{logit}_e\). The magnitude of the resulting change in \(\operatorname{logit}_e\) quantifies the causal effect of \(f_i\) on the final prediction.

\noindent \textbf{Emotion-Selective Causal Sparse Feature Identification}. 
The causal effects of \(f_i \in \mathcal{F}_e\) reveal two types of features: one group exhibits high and relatively uniform effects across all six emotions, while the other group shows substantially stronger effects on specific emotions. We compute the normalized entropy~\citep{shannon48entropy} of each feature based on its causal effect distribution across emotions. Let \(p_e\) denote the normalized absolute causal effect of a feature on emotion \(e\), where
\vspace{-0.5em}
\[
p_e = \frac{|w_e|}{\sum_{e' \in \mathcal{E}} |w_{e'}|}
\]
\vspace{-0.05em}
A low normalized entropy indicates that the feature is highly emotion-selective, while a high normalized entropy suggests that the feature broadly affects multiple emotions. To categorize features based on emotion selectivity, we model the distribution of normalized entropy scores using a Gaussian Mixture Model (GMM)~\citep{Reynolds2009}. Specifically, we fit the GMM over the normalized entropy values of all selected features, allowing the model to separate features into two latent groups with different levels of emotion specificity. Features are assigned to clusters according to their maximum posterior probability under the fitted mixture model. We interpret the low-entropy cluster as emotion-selective features.

To further identify a minimal set of emotion features with the strongest causal effects, we rank features by their individual causal effect estimates and perform top-\(k\) joint feature ablations to measure their combined effect. We experiment with \(k \in \{1, 2, 3, 4, 5, 6, 7, 8, 10, 15, 30\}\), and select the minimum \(k\) at which the causal effect reaches a plateau. We denote the resulting minimal emotion-selective feature set as \(\mathcal{F}_{e,s}\).

\noindent \textbf{Results: Causal Sparse Emotion Features Emerge in Mid-Late Layers}. Across models, we find that a small set of sparse features, averaging 251 features per model and 3.3 features per layer per emotion, accounts for a substantial portion of the causal effect on target emotion logits. These features typically begin to emerge in early mid-to-late layers with relatively weak effects, and their causal influence increases in later layers. This pattern suggests that emotion information is gradually consolidated into sparse, behaviorally relevant features as the model approaches the final prediction. Detailed statistics for causal sparse emotion features are provided in Appendix Table~\ref{tab:feature_summary_by_model_emotion}.

\noindent \textbf{Emotion Features Differ in Concentration Across Emotions}. 
The sparse causal organization of emotion features differs substantially across emotions, as shown in Figure~\ref{fig:figure1}(A). \textcolor{camera}{To characterize this variation, we define \emph{feature concentration} based jointly on the number of emotion-selective causal features and their mean absolute causal effect per feature. A representation is more concentrated when fewer features carry stronger average causal effects. The emotion \textit{fear} exhibits the highest feature concentration, relying on the fewest features per layer with the strongest average causal effect} on \(\operatorname{logit}_{\mathrm{fear}}\), followed by \emph{surprise}. In contrast, \emph{disgust} exhibits the lowest feature concentration, indicating that its causal representation is the most distributed. It relies on more causal features, each of which tends to have a weaker effect on \(\operatorname{logit}_{\mathrm{disgust}}\). \emph{Sadness} shows a similar, though less extreme, distributed pattern.


\noindent \textbf{Emotion Features Are Not Fully Separated}.
\textcolor{camera}{We observe that some causal sparse features have substantial effects on multiple emotion logits under ablation. We refer to this sharing as \emph{feature overlap} and quantify it using the Jaccard index (JI) between the selected feature sets of two emotions. A higher JI indicates less separation between the emotions in the SAE feature space. Across models, the highest mean JI occurs between \emph{disgust} and \emph{anger}, with a mean of \(0.12\) (Figure~\ref{fig:figure1}(B)). This overlap is most pronounced in Gemma-2-2B (\(\mathrm{JI}=0.19\)) and less pronounced among the Gemma-3-12B variants, which have a mean JI of \(0.065\) (see Appendix Figure~\ref{fig:appendix_all_models_jaccard_overlap}). We also observe moderate overlap between \emph{surprise} and \emph{joy}, with a mean JI of \(0.05\).}

\noindent \textbf{Activation-Strength Correlations Reveal Potential Sources of Emotion Confusion}.
\textcolor{camera}{In addition to \emph{feature overlap}, different emotion feature sets can activate together on the same instances, as shown in Figure~\ref{fig:figure1}(C). For example, \emph{surprise} features may activate alongside \emph{joy} features on an instance labeled with \emph{joy}. We refer to this behavior as \emph{feature co-activation} and quantify it using the Pearson correlation between the mean activation strengths of two emotion feature sets across instances. Across models, the mean Pearson correlations are positive between \emph{joy} and \emph{surprise} features (\(\bar{r} = 0.30\)) and between \emph{disgust} and \emph{anger} features (\(\bar{r} = 0.48\)). These correlations suggest that models tend to activate related emotion feature sets with similar strength on the same instances, potentially contributing to emotion confusion. Additional model-level correlation results appear in Appendix Figure~\ref{fig:appendix_all_models_activation_strength_corr}.}

\noindent \textbf{Distributed Emotions May Contain Multiple Semantic Substructures.} 
Finally, we ask whether distributed emotions are represented by many redundant weak features or by features that respond to distinct semantic sub-concepts. For the more distributed emotions, \emph{joy}, \emph{sadness}, and \emph{disgust}, we cluster feature activation patterns over predicted examples using \(k\)-Means clustering and qualitatively inspect the top-activating instances for each cluster. We find that these emotions are not always represented by redundant copies of the same signal. In some models, \emph{joy} separates into social or experiential joy and achievement- or reward-based joy. \emph{Disgust} also shows semantic heterogeneity in some models, sometimes splitting between physical contamination or bodily aversion and social or norm-based aversion, while larger models further divide these into finer subtypes. \emph{Sadness} is less cleanly separable, with most clusters overlapping around a shared loss-centered core involving absence, separation, bereavement, and attachment disruption. Example instances are shown in Appendix Table~\ref{tab:emotion_cluster_examples}.

\section{RQ2: Sparse Emotion Features as Causal Drivers}

\begin{figure*}[t]
    \centering
    \includegraphics[width=\textwidth]{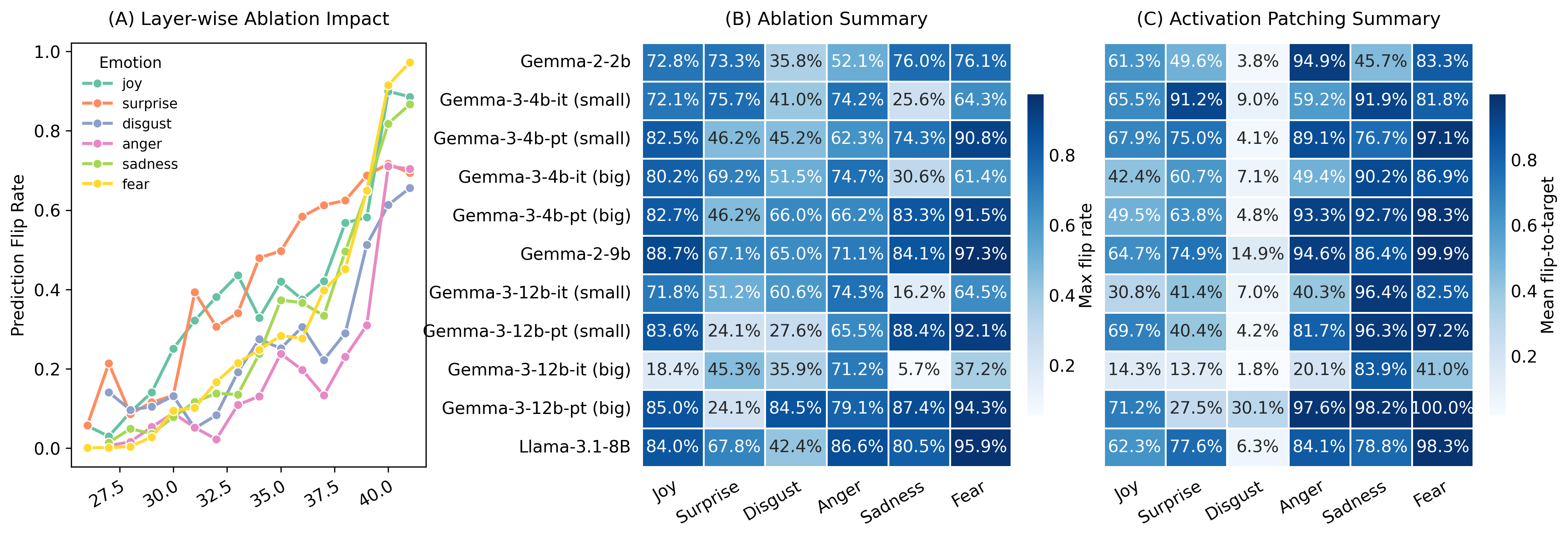}
    \caption{
    Necessity and sufficiency of causal sparse emotion features.
    (A) Layer-wise ablation effects for Gemma-2-9B.
    (B) Maximum flip-away rates across models.
    (C) Maximum flip-to-target rates under activation patching.
    }
    \label{fig:figure2}
\end{figure*}
\vspace{-0.5em}

\noindent \textbf{Necessity and Sufficiency of Sparse Emotion Features}. 
We conduct layer-wise ablation and activation patching experiments to evaluate whether the identified causal sparse emotion features are behaviorally necessary or sufficient for inducing target emotion predictions, and how their causal influence evolves across layers~\citep{meng2022locating, marks2025sparse}. Let \(\mathcal{F}_{e,s}^{(\ell)}\) denote the selected sparse feature set for emotion \(e\) at layer \(\ell\), and let \(z_e(x)\) denote the output logit for emotion \(e\) on input \(x\). For necessity, we ablate all selected features \(f_i^{(\ell)} \in \mathcal{F}_{e,s}^{(\ell)}\) by setting their activations to zero, and compute the flip-away rate as the fraction of originally correct predictions for emotion \(e\) that change to any other emotion after ablation. For sufficiency, we patch the activations of source-emotion features \(f_i^{(\ell)} \in \mathcal{F}_{e_1,s}^{(\ell)}\) from source examples \(x_s \in \mathcal{D}_{e_1}\) into target examples \(x_t \in \mathcal{D}_{e_2}\), and compute the flip-to-target rate as the fraction of patched target examples whose prediction changes to the source emotion \(e_1\). Together, these experiments test how strongly causal sparse emotion features control model predictions, and whether this control varies across layers and emotions.

\noindent \textbf{Results}. In the ablation experiments, we find that the flip rate of target emotion generally increases in deeper layers, as shown for Gemma-2-9B in Figure~\ref{fig:figure2} (A). Ablating the earliest emotion features flips fewer than 20\% of predictions, whereas ablating features in the final layers can increase the flip rate to around 80\%, suggesting that emotion information becomes more concentrated in later layers. Figure~\ref{fig:figure2} (B) shows the variation in maximum flip rate across emotions. \emph{Fear}, which has the most concentrated feature set, shows the highest flip rate, followed by \emph{joy}. In contrast, \emph{disgust} has the lowest flip rate, followed by \emph{surprise}, suggesting that these emotions are less dependent on any single selected feature set and may rely on more distributed causal pathways. Interestingly, the four instruction-tuned Gemma-3 models generally show lower flip rates, especially for \emph{sadness} and \emph{fear}.

Activation patching shows a similar layer-wise trend: flip-to-target rates generally increase in later layers, suggesting that causal sparse emotion features become more behaviorally decisive closer to the output. As shown in Figure~\ref{fig:figure2} (C), we observe clear differences across emotions. \emph{Fear} has the highest flip-to-target rate, followed by \emph{sadness}, indicating that these feature sets are relatively effective at inducing the target emotion prediction. In contrast, \emph{disgust} has the lowest flip-to-target rate, followed by \emph{joy}. \emph{Disgust} features frequently cause predictions to flip to \emph{anger} rather than \emph{disgust}, further suggesting strong entanglement between these two emotions and a comparatively weaker \emph{disgust} representation. Together, these results suggest that emotions differ substantially in how compactly and causally their representations are encoded. \emph{Fear} appears to rely on concentrated and behaviorally effective sparse features, whereas \emph{disgust} is represented more diffusely and remains strongly entangled with causal features for \emph{anger}. Model-level ablation and activation patching results are shown in Appendix Figures~\ref{fig:appendix_all_models_layerwise_ablation} and~\ref{fig:appendix_all_models_layerwise_activation_patching}.

\noindent \textbf{Adjacent Feature Mediation}. 
To examine whether emotion-relevant information is transmitted across sparse features in different layers, we conduct intervention-based restoration experiments between adjacent selected emotion-feature layers in the emotion-selective feature set \(\mathcal{F}_{e,s}\). For each emotion \(e\), we consider adjacent selected layer pairs \(\mathcal{F}_{e,s}^{(\ell)}\) and \(\mathcal{F}_{e,s}^{(m)}\), where \(m\) is the next downstream layer containing selected emotion features. We test whether restoring the downstream feature set \(\mathcal{F}_{e,s}^{(m)}\) can partially recover the target emotion logit after ablating the upstream feature set \(\mathcal{F}_{e,s}^{(\ell)}\). This analysis is inspired by prior work on causal tracing and intervention-based analysis of internal model components~\citep{vig2020causal,meng2022locating}. Formally, let
\vspace{-0.5em}
\[
X = \mathcal{F}_{e,s}^{(\ell)}, \qquad 
M = \mathcal{F}_{e,s}^{(m)}, \qquad 
Y = \operatorname{logit}_e .
\]
\vspace{-0.05em}
We compare four conditions: the clean run, ablating the upstream feature set \(X\), ablating the downstream feature set \(M\), and ablating \(X\) while restoring \(M\) to its clean activation. These interventions test whether upstream emotion features influence downstream emotion features and whether downstream restoration can recover part of the lost emotion signal.
\vspace{-0.4em}
\[
R_{X \to M \to Y}
=
\frac{
s(X{=}0, M{=}\mathrm{clean}) - s(X{=}0)
}{
s(\mathrm{clean}) - s(X{=}0)
},
\]
\vspace{-0.05em}
where \(s(\cdot)\) denotes the emotion logit, \(X{=}0\) denotes upstream feature-set ablation, and \(M{=}\mathrm{clean}\) denotes restoring the downstream feature set to its clean activation. A positive recovery indicates that the downstream sparse feature set carries part of the upstream feature set's task-relevant effect. This analysis tests whether causal sparse emotion features across adjacent layers transmit emotion-relevant information through sequential sparse feature interactions.

\noindent \textbf{Results}. Detailed results are shown in Table~\ref{tab:all_models_adjacent_group_emotion_mean} in the Appendix. Across emotions and models, restoring downstream sparse feature groups often recovers a substantial fraction of the target emotion logit lost after upstream-group ablation. This provides intervention-based evidence that adjacent sparse feature groups carry task-relevant emotion information across layers. Recovery ratios are generally higher for \emph{fear} (73\%--105\%) and lower for \emph{disgust} (51\%--75\%), suggesting differences in how locally recoverable sparse emotion effects are across emotions. Larger Gemma-3 models also tend to show stronger restoration effects overall, although the pattern is consistently observed across model families. These results suggest that sparse emotion features do not operate independently, but instead participate in a sequential pathway of emotion-relevant information across layers.

\section{RQ3: Intervening on Sparse Emotion Features}

In the previous sections, we identified causal sparse emotion features and analyzed their causal mechanisms. We next examine whether these features can be intervened on to control and improve the model's emotion recognition behavior.

\begin{figure*}[t]
    \centering
    \includegraphics[width=0.95\textwidth]{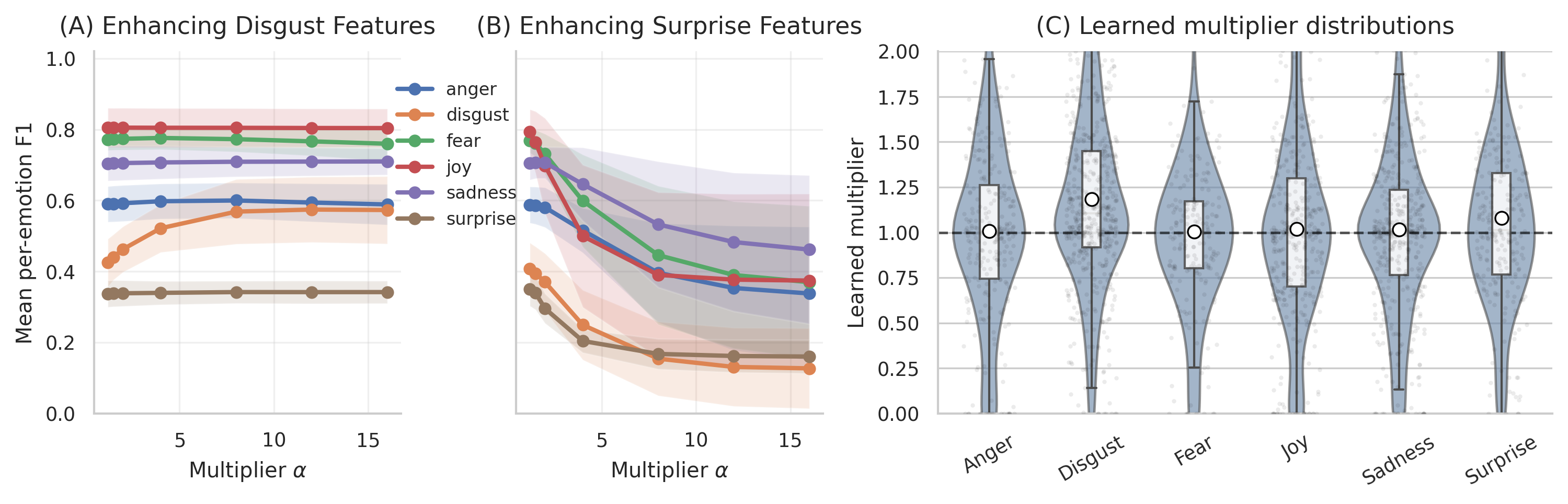}
    \caption{
    Targeted and global sparse feature steering results. 
    (A, B) Aggregate per-emotion F1 across models under increasing feature multiplier \(\alpha\) for targeted \emph{disgust} and \emph{surprise} features. Shaded regions indicate standard deviation across models. 
    (C) Distribution of learned feature multipliers from global sparse feature steering across emotions.}
    \label{fig:figure3}
\end{figure*}

\noindent \textbf{Targeted Sparse Interventions for Weak Emotion Representations}. 
\textcolor{camera}{We conduct a targeted intervention study to test whether naively amplifying identified weak-emotion features can alter and potentially improve recognition performance.} We focus on \emph{disgust} and \emph{surprise}, which show weak baseline performance (see Table~\ref{tab:steering_results}). \textcolor{camera}{For each model and emotion, we select one late layer with an ablation flip rate above 0.2 and amplify all identified causal sparse features for that emotion. We use 0.2 as a heuristic threshold for identifying layers with a substantial intervention effect.} We then amplify their activations by a scalar multiplier \(\alpha \in \{1.2, 1.5, 2, 4, 8, 12, 16\}\) and then evaluate per-emotion F1 scores. Figure~\ref{fig:figure3} (A) shows that increasing \(\alpha\) substantially improves \emph{disgust} F1 from roughly 0.4 to 0.6 before plateauing around \(\alpha=8\), while leaving most other emotions unchanged, suggesting relatively precise controllability. In contrast, Figure~\ref{fig:figure3} (B) shows that naively amplifying \emph{surprise} features broadly degrades emotion prediction performance across nearly all emotions. \textcolor{camera}{On some non-surprise instances, \emph{surprise} features are weakly active alongside more strongly activated features associated with the ground-truth emotion. Amplification increases the \emph{surprise} logit until it can exceed the correct-emotion logit, thereby reducing F1 for other emotions. Additional results for selected model-layer combinations are reported in Appendix Table~\ref{tab:targeted_intervention_results}}.

\noindent \textbf{Global Sparse Feature Steering}. 
Motivated by the targeted intervention results, we next ask whether sparse emotion features can be jointly calibrated to strengthen weak features and improve overall emotion recognition. We conduct a global sparse feature steering experiment in which we learn a feature-wise scaling vector \(S \in \mathcal{R}^d\) over the union of selected causal sparse emotion features, \(\mathcal{F}_s = \bigcup_{e \in \mathcal{E}} \mathcal{F}_{e,s}\).

Importantly, this intervention is highly sparse: instead of updating model weights, we learn only one scalar multiplier for each selected causal sparse feature, resulting in an average of 251 trainable parameters per model. Given the original activations \(F_s\), the steered activations are defined as
\vspace{-0.5em}
\[
F'_s = S \odot F_s,
\]
\vspace{-0.05em}
where \(\odot\) denotes element-wise scaling. We optimize \(S\) using cross-entropy loss for emotion recognition:
\vspace{-0.5em}
\[
L = L_{\mathrm{CE}}(y, \hat{y}).
\]
\vspace{-0.05em}
The training and test sets are drawn from the combined dataset described in the experimental setup. To evaluate whether the learned steering generalizes beyond the training distribution, we further evaluate the steered models on two additional emotion recognition datasets with six basic emotion labels and relatively balanced label distributions: EmoEvent and XED~\cite{plaza-del-arco-etal-2020-emoevent,ohman-etal-2020-xed}. We also compare against few-shot prompting \textcolor{camera}{and emotion-vector steering} as inference-time baselines~\cite{konen-etal-2024-style}. \textcolor{camera}{To test whether the identified features and learned steering generalize across prompt formulations, we repeat the evaluation using three additional prompt templates.} More details are provided in Appendix~\ref{sec:training-details}.

\begin{table*}[t]
\centering
\scriptsize
\setlength{\tabcolsep}{1.5pt}
\renewcommand{\arraystretch}{0.4}
\begin{tabular}{llcccccccccc}
\toprule
\multirow{2}{*}{Model} & \multirow{2}{*}{Setting}
& \multicolumn{7}{c}{Main Dataset (F1)}
& \multirow{2}{*}{EmoEvent}
& \multirow{2}{*}{XED}
& \multirow{2}{*}{PPL} \\
\cmidrule(lr){3-9}
& & Anger & Disgust & Fear & Joy & Sadness & Surprise
& \textbf{Macro} & & & \\
\midrule

\multirow{4}{*}{Gemma-2-2B}
& Zero-shot
& 0.54 & 0.24 & 0.71 & 0.79 & 0.73 & 0.29
& 0.55 & 0.39 & 0.36 & 17.54 \\
& Few-shot
& \textbf{0.64} & 0.61 & \textbf{0.79} & \textbf{0.82}
& 0.75 & 0.34 & 0.66 & 0.37 & 0.42 & -- \\
& Emotion-Vector
& 0.59 & 0.56 & 0.77 & 0.79 & 0.75 & 0.29
& 0.63 & \textbf{0.49} & 0.42 & 17.80 \\
& Steered
& 0.63 & \textbf{0.68} & 0.78 & 0.81 & \textbf{0.76}
& \textbf{0.45} & \textbf{0.69} & \textbf{0.49}
& \textbf{0.49} & 17.62 \\

\midrule
\multirow{4}{*}{Gemma-2-9B}
& Zero-shot
& 0.62 & 0.44 & 0.77 & 0.84 & 0.74 & 0.32
& 0.62 & 0.40 & 0.44 & 13.27 \\
& Few-shot
& 0.66 & 0.58 & 0.80 & 0.79 & \textbf{0.78} & 0.44
& 0.67 & 0.44 & 0.42 & -- \\
& Emotion-Vector
& 0.62 & 0.65 & 0.80 & 0.85 & 0.77 & 0.42
& 0.69 & \textbf{0.49} & 0.50 & 13.66 \\
& Steered
& \textbf{0.67} & \textbf{0.70} & \textbf{0.81}
& \textbf{0.86} & 0.77 & \textbf{0.58}
& \textbf{0.73} & 0.46 & \textbf{0.52} & 13.39 \\

\midrule
\multirow{4}{*}{Gemma-3-12B-IT}
& Zero-shot
& 0.53 & 0.49 & 0.77 & 0.86 & 0.68 & 0.34
& 0.61 & 0.49 & 0.47 & 18.38 \\
& Few-shot
& \textbf{0.68} & 0.58 & \textbf{0.82} & \textbf{0.88}
& \textbf{0.79} & 0.23 & 0.66 & 0.50 & \textbf{0.52} & -- \\
& Emotion-Vector
& 0.61 & 0.66 & 0.77 & 0.86 & 0.76 & 0.34
& 0.67 & \textbf{0.55} & 0.50 & 21.86 \\
& Steered
& 0.66 & \textbf{0.68} & 0.79 & 0.83 & 0.74
& \textbf{0.48} & \textbf{0.70} & 0.51
& 0.46 & 18.31 \\

\midrule
\multirow{4}{*}{Gemma-3-12B-PT}
& Zero-shot
& 0.65 & 0.47 & 0.78 & 0.84 & \textbf{0.79} & 0.37
& 0.65 & 0.43 & 0.43 & 11.19 \\
& Few-shot
& 0.65 & 0.59 & 0.73 & \textbf{0.85} & \textbf{0.79}
& 0.36 & 0.66 & 0.39 & 0.39 & -- \\
& Emotion-Vector
& \textbf{0.67} & 0.68 & \textbf{0.80} & \textbf{0.85}
& \textbf{0.79} & 0.36 & 0.69 & 0.49 & 0.45 & 16.52 \\
& Steered
& 0.65 & \textbf{0.72} & 0.79 & \textbf{0.85} & 0.78
& \textbf{0.46} & \textbf{0.71} & \textbf{0.51}
& \textbf{0.46} & 11.15 \\

\midrule
\multirow{4}{*}{Gemma-3-4B-IT}
& Zero-shot
& 0.54 & 0.42 & 0.74 & 0.83 & 0.65 & 0.27
& 0.58 & 0.46 & 0.45 & 34.26 \\
& Few-shot
& 0.63 & 0.45 & \textbf{0.79} & \textbf{0.84}
& 0.71 & 0.35 & 0.63 & 0.43 & 0.44 & -- \\
& Emotion-Vector
& 0.63 & 0.61 & 0.78 & \textbf{0.84} & 0.72 & 0.34
& 0.65 & \textbf{0.50} & 0.48 & 61.48 \\
& Steered
& \textbf{0.66} & \textbf{0.67} & \textbf{0.79} & 0.82
& \textbf{0.77} & \textbf{0.39} & \textbf{0.68}
& 0.48 & \textbf{0.50} & 32.90 \\

\midrule
\multirow{4}{*}{Gemma-3-4B-PT}
& Zero-shot
& 0.58 & 0.36 & 0.74 & 0.71 & 0.68 & 0.39
& 0.58 & 0.31 & 0.40 & 14.77 \\
& Few-shot
& \textbf{0.66} & 0.53 & \textbf{0.80} & 0.78
& \textbf{0.78} & 0.42 & 0.66 & 0.37
& 0.41 & -- \\
& Emotion-Vector
& 0.63 & 0.57 & 0.77 & 0.78 & 0.74 & 0.37
& 0.64 & \textbf{0.40} & 0.41 & 19.48 \\
& Steered
& 0.64 & \textbf{0.67} & 0.77 & \textbf{0.81} & 0.76
& \textbf{0.46} & \textbf{0.69} & 0.37
& \textbf{0.46} & 14.79 \\

\midrule
\multirow{4}{*}{Llama-3.1-8B}
& Zero-shot
& 0.65 & 0.49 & 0.75 & \textbf{0.84} & \textbf{0.77}
& 0.29 & 0.63 & 0.39 & 0.47 & 11.51 \\
& Few-shot
& \textbf{0.66} & 0.52 & \textbf{0.79} & 0.75
& \textbf{0.77} & 0.39 & 0.65 & 0.35 & 0.41 & -- \\
& Emotion-Vector
& 0.65 & 0.65 & 0.76 & 0.81 & \textbf{0.77} & 0.32
& 0.66 & \textbf{0.46} & 0.49 & 11.52 \\
& Steered
& \textbf{0.66} & \textbf{0.66} & 0.77 & 0.83
& \textbf{0.77} & \textbf{0.44} & \textbf{0.69}
& 0.42 & \textbf{0.51} & 11.53 \\

\bottomrule
\end{tabular}
\caption{Emotion recognition performance across zero-shot, few-shot, emotion-vector steering, and sparse-feature steering settings. For each
model, the best F1 score among available settings is shown in \textbf{bold}.
EmoEvent and XED evaluate generalization to external datasets. Perplexity (PPL)
measures preservation of general language-modeling ability.}
\label{tab:steering_results}
\end{table*}

\noindent \textbf{Results}. Table~\ref{tab:steering_results} reports per-emotion F1 scores, generalization performance on two external datasets, and language modeling perplexity. In the zero-shot setting, models perform particularly poorly on \emph{disgust} and \emph{surprise}: averaged across models, F1 is 0.42 for \emph{disgust} and 0.32 for \emph{surprise}, compared with 0.73--0.82 for \emph{fear}, \emph{joy}, and \emph{sadness}. This is consistent with our earlier analysis that \emph{disgust} is more distributed and weakly represented, while \emph{surprise} features are often co-activated with other emotions, especially \emph{joy}. After steering, average F1 improves from 0.42 to 0.68 for \emph{disgust} (64.3\% relative increase) and from 0.32 to 0.47 for \emph{surprise} (43.6\%), while macro-F1 increases from 0.60 to 0.70 on average across models (15.9\%). These improvements also transfer to external datasets, with average macro-F1 increasing from 0.41 to 0.46 on EmoEvent (12.9\%) and from 0.43 to 0.49 on XED (12.6\%). These improvements are significant under paired Wilcoxon signed-rank tests (\(p < 0.05\)); see Appendix Section~\ref{sec:statistic}. Perplexity remains largely unchanged. Few-shot prompting also improves main-dataset performance, but has smaller gains on weaker emotions and transfers less consistently. \textcolor{camera}{Compared with sparse-feature steering, emotion-vector steering yields a smaller average macro-F1 improvement on the main dataset, with especially limited gains for \emph{surprise}, and substantially increases perplexity for some models.}

Figure~\ref{fig:figure3} (C) shows the aggregated distributions of learned multipliers for emotion-related features across all models. We find that the learned multipliers are generally centered around 1.0, indicating that the optimization preserves most feature activations close to their original scale rather than applying extreme steering. However, the distributions differ across emotions: \emph{disgust} shows a noticeably higher median multiplier, while \emph{surprise} shows a slightly higher median multiplier, suggesting that features associated with these emotions often require stronger amplification. Together, these results show that sparse emotion features provide an effective intervention target for improving emotion recognition.

\noindent \textbf{Lexical-Shortcut and Label-Prior Controls}. 
\textcolor{camera}{To test whether the observed steering gains are driven by label-correlated lexical shortcuts or evaluation-set label imbalance, we conduct an additional controlled evaluation. We identify highly label-correlated unigrams and bigrams, remove examples containing these cues, and balance the remaining examples across emotion labels. We then evaluate the filtered dataset using the same zero-shot prompt and the steering parameters learned on the original training set, without retraining. As shown in Table~\ref{tab:lexical_shortcut_control}, the steered models consistently outperform their unsteered counterparts, improving average macro-F1 from \(0.54\) to \(0.65\). This \(0.11\) absolute improvement is comparable to that observed on the unfiltered dataset. These results provide additional evidence that the steering gains cannot be fully explained by the identified \(n\)-gram shortcuts or label imbalance alone. Although this experiment does not fully establish the sources of emotion-recognition difficulty, it strengthens the interpretation that the identified sparse features capture causally relevant aspects of the models' internal emotion representations.}

\section{Conclusion}

Large language models are increasingly used in emotionally sensitive human-AI interactions, where reliable emotion recognition is important for safety. In this work, we study emotion-specific biases in LLMs through causal sparse emotion features identified with SAEs. We show that emotions differ substantially in their sparse causal organization. Some emotions, such as \emph{fear} and \emph{surprise}, are supported by more concentrated causal feature sets, while \emph{disgust} is represented by weaker and more distributed features. We further show that emotion features are not fully separated: \emph{disgust} overlaps and covaries strongly with \emph{anger}, while \emph{surprise} is closely associated with \emph{joy}. Through ablation and activation patching, we find that these features become more behaviorally decisive in later layers, but their necessity and sufficiency vary substantially across emotions. Restoration experiments further suggest that sparse emotion features participate in a sequential flow of emotion-relevant information across layers. Together, these results provide a mechanistic explanation for why models struggle more with certain emotions and why specific confusion patterns emerge.

We further examine whether causal sparse emotion features can serve as intervention targets. Our steering experiments show that feature-level calibration can strengthen weakly represented emotions, improve emotion recognition across models, transfer to external datasets, and largely preserve language modeling ability.

Overall, our findings show that emotion recognition failures in LLMs are not merely behavioral errors, but reflect differences in the internal sparse causal organization of emotion representations. By identifying and intervening on causal sparse emotion features, this work provides an interpretable path for diagnosing and improving affective capabilities in LLMs. Future work can investigate why certain emotion feature sets co-activate from input cues and why some emotion features become weaker during training. \textcolor{camera}{The code and dataset are available at \url{https://github.com/Computing-for-Social-Good-CSG/sae-emotion}}

\section*{Limitations}

Due to the limited availability of publicly released full-layer SAEs from GemmaScope, GemmaScope 2, and LlamaScope, we restrict our analysis to seven models. Although these models span different scales, model families, and instruction-tuning settings, most are from the Gemma family. Other available SAE releases are less suitable for our study: some are trained on models that are too small to reliably perform emotion inference, such as GPT-2 and Pythia-70M~\cite{karvonen2024evaluatingsparseautoencoderstargeted,gao2025scaling}, while others do not provide SAEs for all layers, such as GPT-OSS-20B SAE~\cite{goodfire2025hackathon_gptoss20bSAE}. 

In addition, although we observe similar trends between SAEs with widths of 16k and 32k, we do not explore substantially wider SAEs (e.g., 161k features) in order to maintain consistency across models. Wider SAEs may further reduce polysemanticity and capture more fine-grained concepts, which could lead to a more detailed characterization of emotion representations.

\textcolor{camera}{Our analysis is further restricted to six basic emotions under a single-label formulation. This controlled setting facilitates consistent SAE feature analysis across datasets with different emotion-label inventories, but simplifies the ambiguity and co-occurrence inherent in emotional experience. Although our datasets use self-reported emotion labels, reducing reliance on third-party interpretation, self-reports do not eliminate overlapping affective states or unclear category boundaries~\cite{mauss2009measures,schuff-etal-2017-annotation}. For example, an instance labeled as \emph{disgust} may also contain secondary \emph{anger}; consequently, feature overlap and co-activation may partly reflect genuine emotional overlap rather than model error. The consistent tendency of the evaluated models to favor \emph{anger} over \emph{disgust} suggests an additional model-specific asymmetry, but our experiments cannot fully disentangle learned representational organization from ambiguity in the emotion labels. Future work should extend this framework to fine-grained, compound, and multi-label emotions.}

\section*{Ethical Considerations}

Our study provides evidence of how large language models encode and process emotion-related information and identifies features that influence emotion predictions. However, these findings do not imply that LLMs possess human-like emotional reasoning abilities. Rather, the mechanisms identified in our analysis reflect statistical patterns in model representations. Emotional intelligence (EI) in LLMs remains an open research problem, and there is currently no widely agreed-upon definition of what constitutes safe or reliable emotional behavior in such systems.

While our method improves emotion recognition performance, improved performance on benchmark datasets does not guarantee safety or reliability in real-world applications. Emotion recognition in LLMs reflects learned patterns from data, which may be incomplete, biased, or culturally dependent. We observe uneven performance across emotions (e.g., weaker representation of \emph{disgust}), which raises concerns about consistency and fairness. Misclassification of user emotions in real-world interactions, particularly in sensitive domains such as mental health support, may lead to inappropriate or harmful responses, potentially affecting vulnerable users~\cite{ball2001emotion}. Moreover, interpretability methods may create a false sense of transparency and lead users to overestimate the model’s understanding or reliability, as such features are approximations rather than complete causal explanations.

Our steering approach introduces additional considerations. Although it enables targeted modification of model behavior, it also presents dual-use risks, as similar techniques could be used to manipulate emotional interpretations or influence user behavior in undesirable ways. For these reasons, we emphasize that our work is intended to advance scientific understanding of model mechanisms rather than enable direct deployment in high-stakes or human-facing applications. Any real-world use should be approached with caution and accompanied by appropriate safeguards, evaluation, and human oversight.


\bibliography{anthology,custom}

\appendix

\renewcommand{\thefigure}{A.\arabic{figure}}
\renewcommand{\thetable}{A.\arabic{table}}
\setcounter{figure}{0}
\setcounter{table}{0}

\section{Appendix}
\label{sec:appendix}

\subsection{Sparse Autoencoders}
\label{sec:sparse-autoencoders}

Neural networks represent information in high-dimensional activation spaces, but the number of concepts a model must encode far exceeds the number of available neurons. This leads to \textit{superposition}, where multiple concepts are simultaneously encoded as overlapping linear combinations of neuron activations \cite{elhage2022toymodelssuperposition}. As a result, individual neurons are typically \textit{polysemantic}, activating for multiple unrelated concepts, making the direct interpretation of neuron activations unreliable \cite{elhage2022toymodelssuperposition, arora-etal-2018-linear}.

Sparse autoencoders (SAEs) are trained to decompose these polysemantic activations into a larger set of approximately monosemantic features \cite{cunningham2023sparseautoencodershighlyinterpretable, bricken2023monosemanticity}. Formally, an SAE consists of an encoder $f_{\text{enc}}$ and a decoder $f_{\text{dec}}$. Given a hidden state $h^{(\ell)} \in \mathcal{R}^{d_{\text{model}}}$ from layer $\ell$, the encoder maps it to a sparse latent representation:

\begin{equation}
    z^{(\ell)} = \text{ReLU}(W_{\text{enc}}\, h^{(\ell)} + b_{\text{enc}})
\end{equation}

where $W_{\text{enc}} \in \mathcal{R}^{d_{\text{SAE}} \times d_{\text{model}}}$, $d_{\text{SAE}} \gg d_{\text{model}}$, and the ReLU activation enforces sparsity by zeroing out weakly activated features. The decoder reconstructs the original activation as:

\begin{equation}
    \hat{h}^{(\ell)} = W_{\text{dec}}\, z^{(\ell)} + b_{\text{dec}}
\end{equation}

The SAE is trained to minimize a combination of reconstruction error and an $\ell_1$ sparsity penalty:

\begin{equation}
    \mathcal{L}_{\text{SAE}} = \|h^{(\ell)} - \hat{h}^{(\ell)}\|_2^2 + \beta \|z^{(\ell)}\|_1
\end{equation}

where $\beta$ controls the sparsity--reconstruction tradeoff. The $\ell_1$ penalty encourages each input to activate only a small number of latent dimensions, enabling improved interpretability and concept localization \cite{bricken2023monosemanticity}. Each dimension of $z^{(\ell)}$ corresponds to a learned feature direction in the residual stream, and the columns of $W_{\text{dec}}$ define the corresponding feature vectors. Since $d_{\text{SAE}} \gg d_{\text{model}}$, the SAE can represent far more features than there are neurons, providing a richer and more disentangled foundation for interpreting a model's internal representations \cite{cunningham2023sparseautoencodershighlyinterpretable, templeton2024scaling}.

Some SAE variants replace the standard ReLU activation with a learned threshold, known as a JumpReLU, which activates a feature only when its pre-activation exceeds a learnable per-feature threshold $\theta$:

\begin{equation}
    z^{(\ell)} = \text{JumpReLU}_{\theta}(W_{\text{enc}}\, h^{(\ell)} + b_{\text{enc}})
\end{equation}

This formulation allows the model to more precisely control the sparsity level of each feature independently, yielding more faithful reconstructions under the same sparsity constraints \cite{rajamanoharan2024improvingdictionarylearninggated}. The SAEs provided by GemmaScope, which we use in this work, employ this JumpReLU architecture \cite{lieberum2024gemmascopeopensparse}.

\subsection{Data Preprocessing}
\label{sec:appendix-data-prep}

Among these datasets, EXPRESS differs from the others in several important ways. It contains instances with a wide range of text lengths and allows one or multiple emotion labels. In addition, the emotion label is masked from the original text, meaning that the annotated emotion may correspond only to a localized emotional expression rather than the overall emotional content. To ensure consistency across datasets, we apply the following filtering steps to EXPRESS: (1) excluding instances exceeding a fixed length threshold of 50 words, (2) retaining only instances annotated with a single emotion, and (3) using GPT-4o to filter out cases where the labeled emotion reflects only a local emotional span rather than the global emotional tone of the text.

\begin{table*}[t]
\centering
\scriptsize
\setlength{\tabcolsep}{2pt}
\renewcommand{\arraystretch}{0.75}
\begin{tabular}{lccccccccccccc}
\toprule
\multirow{2}{*}{Model} & \multirow{2}{*}{Total}
& \multicolumn{6}{c}{Number of Features}
& \multicolumn{6}{c}{Average Logit Effect per Feature} \\
\cmidrule(lr){3-8} \cmidrule(lr){9-14}
& & Anger & Disgust & Fear & Joy & Sadness & Surprise
& Anger & Disgust & Fear & Joy & Sadness & Surprise \\
\midrule
Gemma-2-2B & 121 & 13 & 38 & 10 & 28 & 33 & 18 & 0.564 & 0.155 & 1.354 & 0.346 & 0.236 & 0.873 \\
Gemma-2-9B & 264 & 35 & 86 & 26 & 57 & 67 & 35 & 0.287 & 0.109 & 0.704 & 0.220 & 0.154 & 0.718 \\
Gemma-3-4B-PT-big & 313 & 62 & 114 & 30 & 55 & 70 & 44 & 0.226 & 0.144 & 0.638 & 0.260 & 0.220 & 0.471 \\
Gemma-3-4B-PT-small & 134 & 25 & 36 & 14 & 30 & 38 & 19 & 0.432 & 0.256 & 1.274 & 0.551 & 0.330 & 1.154 \\
Gemma-3-4B-IT-big & 322 & 47 & 126 & 39 & 68 & 58 & 48 & 0.695 & 0.225 & 0.718 & 0.548 & 0.442 & 0.753 \\
Gemma-3-4B-IT-small & 186 & 30 & 64 & 33 & 33 & 55 & 22 & 1.099 & 0.430 & 1.002 & 1.166 & 0.442 & 1.875 \\
Gemma-3-12B-PT-big & 437 & 71 & 109 & 58 & 109 & 106 & 52 & 0.285 & 0.159 & 0.422 & 0.204 & 0.161 & 0.663 \\
Gemma-3-12B-PT-small & 170 & 29 & 37 & 22 & 40 & 31 & 19 & 0.499 & 0.284 & 1.097 & 0.518 & 0.450 & 1.684 \\
Gemma-3-12B-IT-big & 518 & 75 & 112 & 103 & 135 & 98 & 74 & 0.340 & 0.291 & 0.202 & 0.219 & 0.238 & 0.459 \\
Gemma-3-12B-IT-small & 161 & 26 & 34 & 20 & 34 & 26 & 23 & 1.104 & 0.664 & 2.073 & 0.901 & 1.026 & 2.001 \\
Llama-3.1-8B & 138 & 24 & 30 & 15 & 28 & 37 & 21 & 0.662 & 0.326 & 1.954 & 0.638 & 0.453 & 1.433 \\
\midrule
Average & 251.3 & 39.7 & 71.5 & 33.6 & 56.1 & 56.3 & 34.1 & 0.563 & 0.277 & 1.040 & 0.507 & 0.378 & 1.098 \\
\bottomrule
\end{tabular}
\caption{Summary of selected causal sparse emotion features across models and SAE settings. For each model, we report the total number of unique emotion-selected features, the number of emotion-selective features per emotion, and the average logit effect per feature for each emotion. The final row reports the average across model-SAE settings.}
\label{tab:feature_summary_by_model_emotion}
\end{table*}

\begin{table*}[t]
    \centering
    \scriptsize
    \setlength{\tabcolsep}{6pt}
    \begin{tabular}{llrccc}
        \toprule
        \textbf{Model} & \textbf{Target} & \textbf{Layer}
        & \multicolumn{3}{c}{\textbf{F1 (target / other)}} \\
        \cmidrule(lr){4-6}
        & & & \textbf{Base} & \(\boldsymbol{\alpha=2}\) & \(\boldsymbol{\alpha=8}\) \\
        \midrule
        Gemma-2-2B   & Disgust  & 20 & 0.29 / 0.56 & 0.32 / 0.56 & 0.51 / 0.59 \\
        Gemma-2-2B   & Disgust  & 24 & 0.29 / 0.56 & 0.36 / 0.57 & 0.62 / 0.60 \\
        Gemma-2-2B   & Surprise & 20 & 0.30 / 0.56 & 0.30 / 0.55 & 0.13 / 0.22 \\
        Gemma-2-2B   & Surprise & 24 & 0.30 / 0.56 & 0.32 / 0.56 & 0.22 / 0.48 \\
        \midrule
        Gemma-2-9B   & Disgust  & 34 & 0.44 / 0.63 & 0.47 / 0.64 & 0.69 / 0.68 \\
        Gemma-2-9B   & Disgust  & 37 & 0.44 / 0.63 & 0.49 / 0.65 & 0.68 / 0.68 \\
        Gemma-2-9B   & Surprise & 34 & 0.34 / 0.63 & 0.38 / 0.64 & 0.21 / 0.48 \\
        Gemma-2-9B   & Surprise & 37 & 0.34 / 0.63 & 0.37 / 0.63 & 0.14 / 0.32 \\
        \midrule
        Llama-3.1-8B & Disgust  & 27 & 0.49 / 0.63 & 0.55 / 0.65 & 0.62 / 0.65 \\
        Llama-3.1-8B & Disgust  & 30 & 0.49 / 0.63 & 0.51 / 0.64 & 0.29 / 0.59 \\
        Llama-3.1-8B & Surprise & 27 & 0.29 / 0.63 & 0.32 / 0.62 & 0.18 / 0.31 \\
        Llama-3.1-8B & Surprise & 30 & 0.29 / 0.63 & 0.36 / 0.63 & 0.23 / 0.51 \\
        \bottomrule
    \end{tabular}
    \caption{Sampled feature-boosting results by target emotion and intervention layer. Each entry reports target-emotion F1 / other-emotion F1.}
    \label{tab:targeted_intervention_results}
\end{table*}

\begin{figure*}[t]
    \centering
    \includegraphics[width=\textwidth]{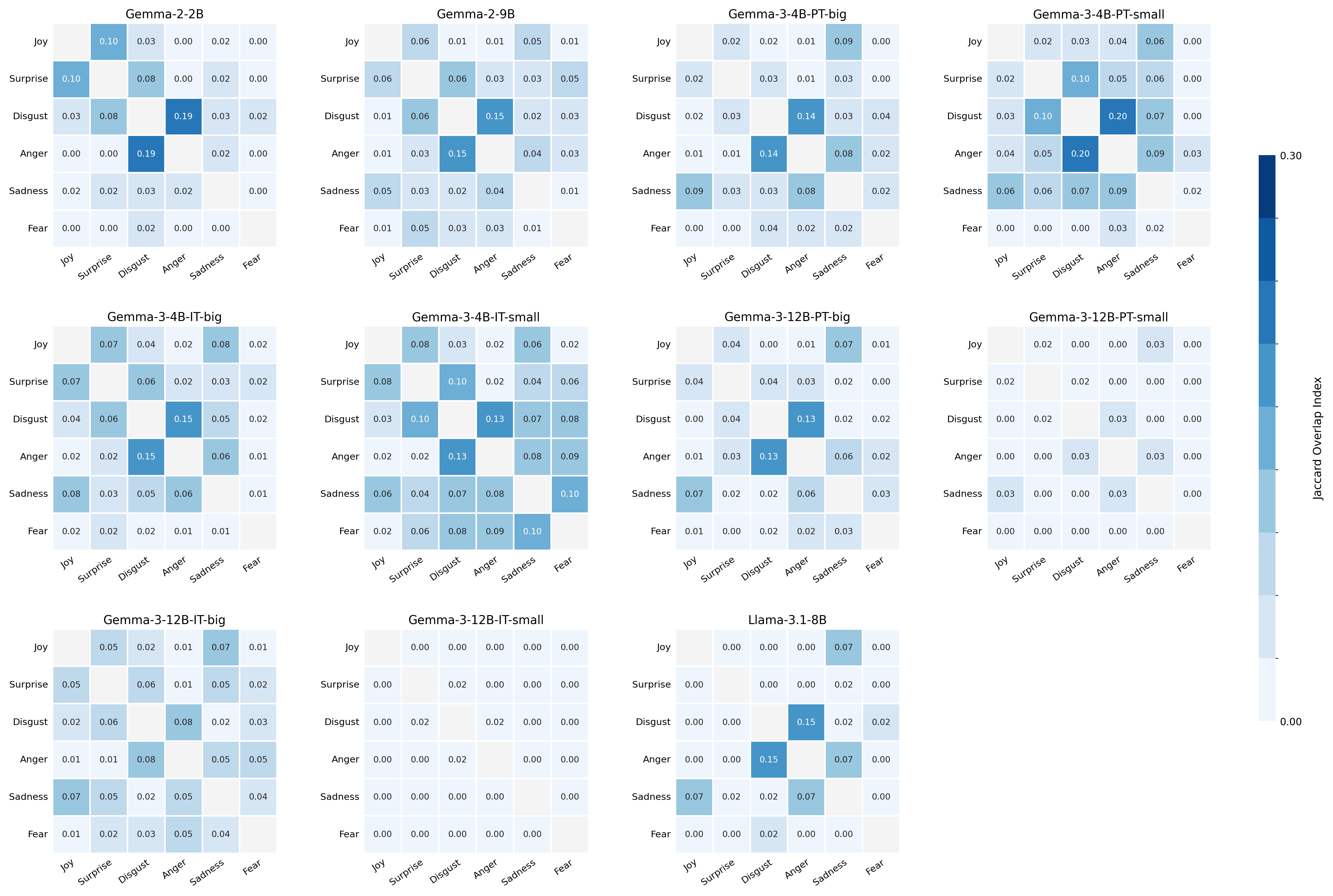}
    \caption{
    Cross-emotion overlap of causal sparse emotion features across all models and SAE settings. 
    Each subplot shows the Jaccard overlap between selected feature sets for each pair of emotions. 
    Diagonal entries are masked, and darker cells indicate stronger feature sharing.}
    \label{fig:appendix_all_models_jaccard_overlap}
\end{figure*}

\begin{figure*}[t]
    \centering
    \includegraphics[width=\textwidth]{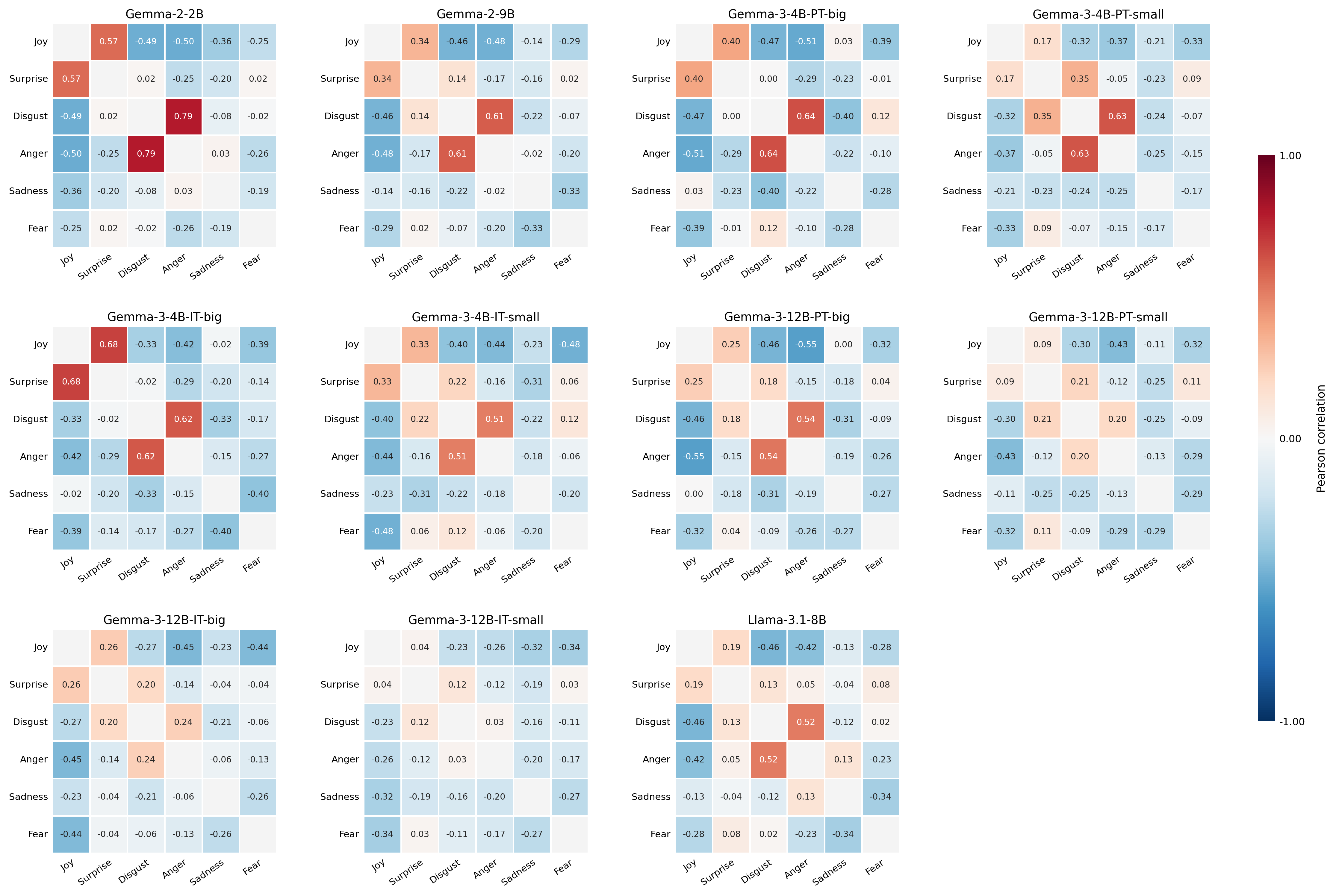}
    \caption{
    Cross-emotion activation-strength correlations across all models and SAE settings.
    Each subplot shows Pearson correlations between the mean activation strengths of causal sparse feature sets for different emotions.
    Diagonal entries are masked. Redder cells indicate stronger positive correlations, while bluer cells indicate stronger negative correlations.
    }
    \label{fig:appendix_all_models_activation_strength_corr}
\end{figure*}

\begin{figure*}[t]
    \centering
    \includegraphics[width=\textwidth]{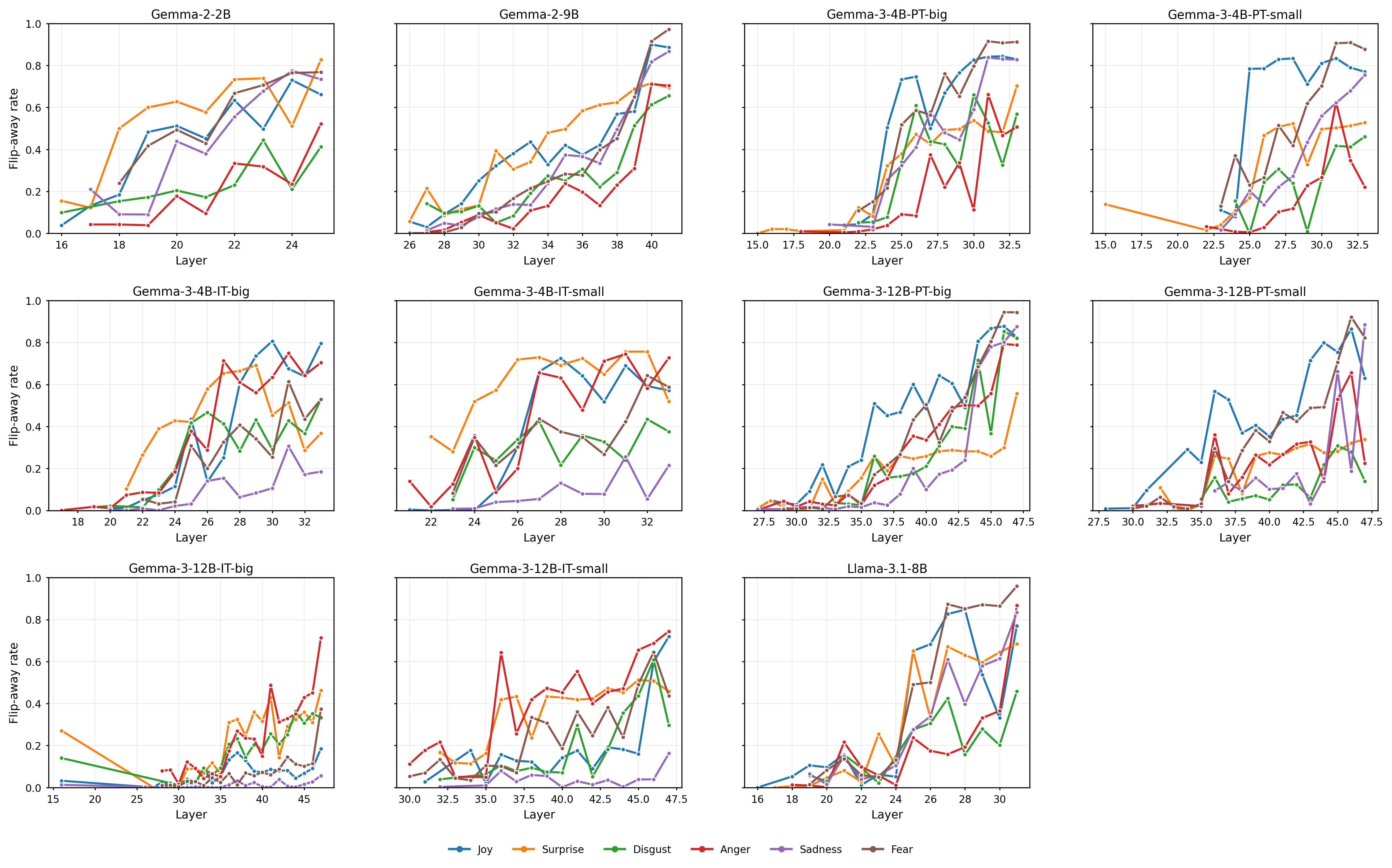}
    \caption{
    Layer-wise ablation effects across all models and SAE settings.
    Each subplot shows the prediction flip-away rate after ablating selected causal sparse emotion features for each emotion across layers.
    Higher values indicate stronger behavioral necessity of the selected features at that layer.
    }
    \label{fig:appendix_all_models_layerwise_ablation}
\end{figure*}

\begin{figure*}[t]
    \centering
    \includegraphics[width=\textwidth]{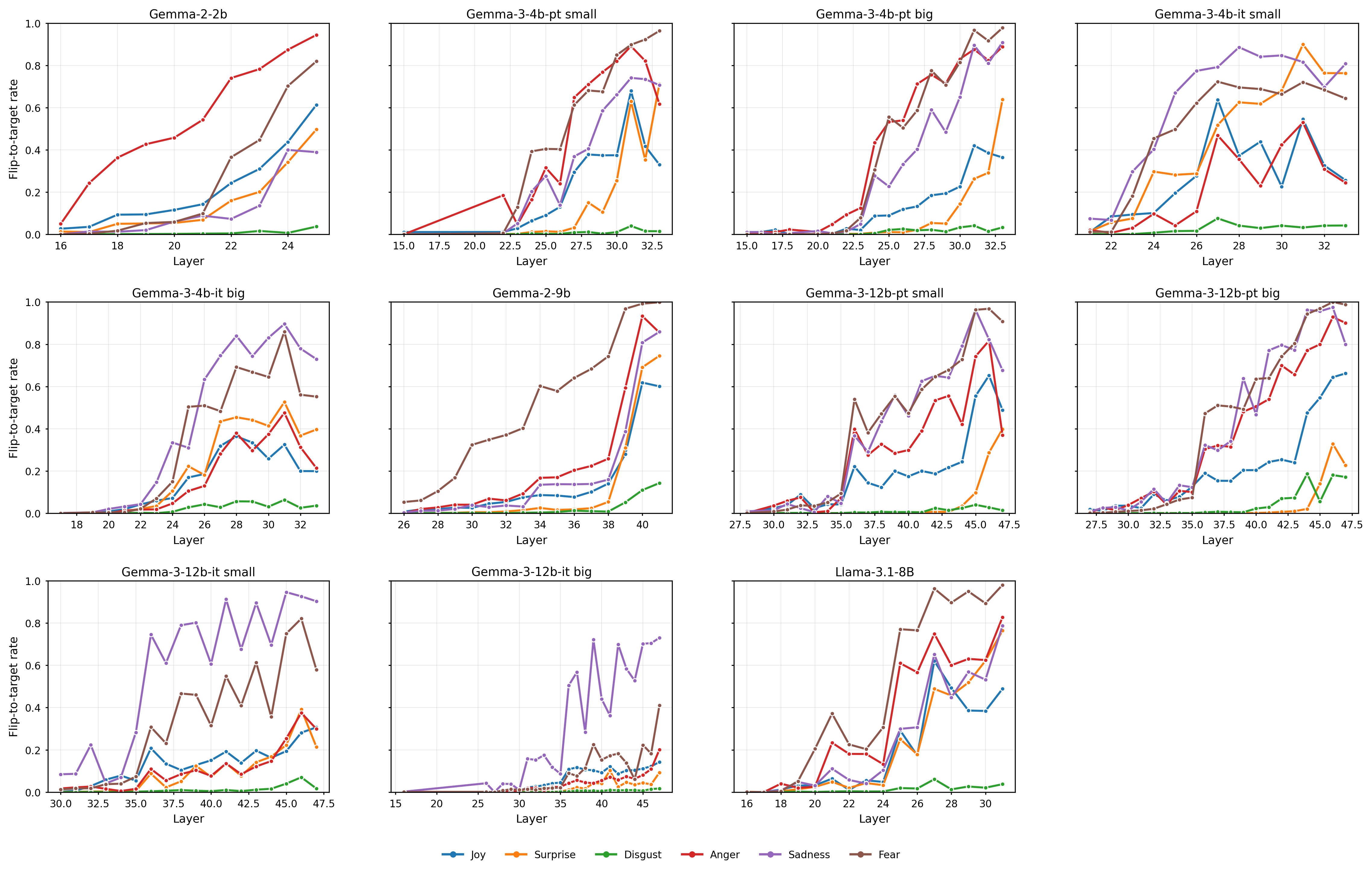}
    \caption{
    Layer-wise activation patching effects across all models and SAE settings.
    Each subplot shows the flip-to-target rate after patching selected causal sparse emotion features for each target emotion across layers.
    Higher values indicate stronger sufficiency of the selected features for inducing the target emotion prediction.
    }
    \label{fig:appendix_all_models_layerwise_activation_patching}
\end{figure*}

\subsection{Global Steering Details}
\label{sec:training-details}

\noindent \textbf{Model Training Details}. 
In the main steering experiment, we use an 80--20 train--test split on the main dataset and sample 400 training examples per emotion from the training set. For Gemma-3 models, we use the small-sparsity SAE setting for steering.

\noindent \textbf{Emotion-Vector Steering Baseline}. 
\textcolor{camera}{We construct a contrastive emotion vector for each emotion \(e\) from the difference between the mean hidden representations of examples associated with that emotion and those associated with the remaining emotions:}
\[
\mathbf{v}_e =
\operatorname{normalize}\!\left(
\mathbb{E}[\mathbf{h}\mid y=e]
-
\mathbb{E}[\mathbf{h}\mid y\neq e]
\right).
\]
\textcolor{camera}{where \(\mathbf{h}\) denotes the hidden representation at a given layer. We then steer the model by learning an emotion-specific scaling factor for each vector while keeping the model parameters frozen. For each model, we perform a layer sweep and report results from the layer that achieves the best validation performance.}

\noindent \textbf{Robustness to Prompt Variation}.
\textcolor{camera}{To evaluate whether the steering effects of the identified emotion-selective features generalize across prompt formulations, we use the three additional semantically equivalent prompt templates shown in Table~\ref{tab:prompt_templates}, adapted from prior work~\citep{tak-etal-2025-mechanistic}.}

\textcolor{camera}{We evaluate the models steered under the original prompt using each of the three additional prompt templates and compare their performance with that of the corresponding vanilla models. Table~\ref{tab:prompt_robustness} reports the per-emotion and macro-F1 scores averaged across these prompts. Steering consistently improves recognition of the two weaker emotions, \emph{disgust} and \emph{surprise}, across all seven models. Macro-F1 also improves for six of the seven models, with Gemma-3-12B-IT as the only exception. These results indicate that the improvements obtained by steering the selected emotion features are robust to reasonable variations in prompt phrasing.}

\begin{table}[!t]
\centering
\footnotesize
\setlength{\tabcolsep}{4pt}
\begin{tabular}{@{}lp{0.80\columnwidth}@{}}
\toprule
Prompt & Template \\
\midrule
P2 &
\raggedright
\texttt{What is the inferred emotion in the following context?}\newline
\texttt{Context: \{text\}.}\newline
\texttt{Answer:}
\tabularnewline
\addlinespace[2pt]
P3 &
\raggedright
\texttt{Consider this list of emotions: anger, disgust, fear, joy, sadness, surprise.}\newline
\texttt{What is the inferred emotion in the following context?}\newline
\texttt{Context: \{text\}.}\newline
\texttt{Answer:}
\tabularnewline
\addlinespace[2pt]
P4 &
\raggedright
\texttt{Guess the emotion.}\newline
\texttt{Context: \{text\}.}\newline
\texttt{Answer:}
\tabularnewline
\bottomrule
\end{tabular}
\caption{Additional prompt templates used in the prompt-variation robustness experiment. \texttt{\{text\}} denotes the input context.}
\label{tab:prompt_templates}
\end{table}

\noindent \textbf{General-Capability Preservation}. 
\textcolor{camera}{Sparse feature steering targets a small set of approximately monosemantic features and may therefore produce fewer off-target effects than dense activation steering. We assess language-modeling preservation using perplexity on a subset of the WikiText-103 test set~\cite{merity2017pointer}. We further evaluate the vanilla and steered models on four complementary benchmarks: MMLU, which measures knowledge and problem-solving across 57 academic subjects~\cite{hendryckstest2021}; ARC-Challenge, which evaluates reasoning over challenging grade-school science questions~\cite{clark2018thinksolvedquestionanswering}; BoolQ, which evaluates reading comprehension through naturally occurring yes-or-no questions~\cite{clark-etal-2019-boolq}; and PIQA, which evaluates physical commonsense reasoning in everyday situations~\cite{bisk2020piqa}. As shown in Table~\ref{tab:general_capability_preservation}, performance remains nearly unchanged after steering. The mean score across the four benchmarks and seven models is \(0.74\) for both the vanilla and steered settings, and no individual model-benchmark score changes by more than \(0.01\). These results provide additional evidence that sparse feature steering preserves the models' general capabilities on the evaluated tasks.}

\begin{table}[!t]
\centering
\small
\setlength{\tabcolsep}{4pt}
\begin{tabular}{lrrrr}
\toprule
\textbf{Model}
& \(\boldsymbol{n}\)
& \textbf{Vanilla}
& \textbf{Steered}
& \(\boldsymbol{\Delta}\) \textbf{F1} \\
\midrule
Gemma-2-2B    & 1332 & 0.49 & \textbf{0.62} & +0.13 \\
Gemma-2-9B    & 1332 & 0.56 & \textbf{0.68} & +0.12 \\
Gemma-3-12B-IT & 1332 & 0.56 & \textbf{0.65} & +0.09 \\
Gemma-3-12B-PT & 1332 & 0.59 & \textbf{0.67} & +0.08 \\
Gemma-3-4B-IT  & 1332 & 0.53 & \textbf{0.64} & +0.11 \\
Gemma-3-4B-PT  & 1332 & 0.51 & \textbf{0.62} & +0.11 \\
Llama-3.1-8B   & 1332 & 0.56 & \textbf{0.65} & +0.09 \\
\bottomrule
\end{tabular}
\caption{Macro-F1 on the filtered and label-balanced dataset after removing examples containing highly label-correlated unigrams and bigrams. The original steering parameters are held fixed. \(n\) denotes the number of retained
evaluation examples.}
\label{tab:lexical_shortcut_control}
\end{table}

\begin{table*}[t]
\centering
\small
\setlength{\tabcolsep}{4pt}
\renewcommand{\arraystretch}{0.85}
\begin{tabular}{llccccccc}
\toprule
\multirow{2}{*}{Model}
& \multirow{2}{*}{Setting}
& \multicolumn{7}{c}{F1 averaged across additional prompts} \\
\cmidrule(lr){3-9}
& & Anger & Disgust & Fear & Joy & Sadness & Surprise & Macro \\
\midrule

\multirow{2}{*}{Gemma-2-2B}
& Vanilla
& 0.50 & 0.26 & 0.62 & 0.75 & 0.55 & 0.25 & 0.49 \\
& Steered
& 0.54 & 0.64 & 0.74 & 0.78 & 0.69 & 0.28 & 0.61 \\

\midrule
\multirow{2}{*}{Gemma-2-9B}
& Vanilla
& 0.65 & 0.51 & 0.80 & 0.86 & 0.77 & 0.33 & 0.65 \\
& Steered
& 0.68 & 0.71 & 0.82 & 0.75 & 0.78 & 0.45 & 0.70 \\

\midrule
\multirow{2}{*}{Gemma-3-4B-PT}
& Vanilla
& 0.60 & 0.42 & 0.74 & 0.78 & 0.72 & 0.27 & 0.59 \\
& Steered
& 0.59 & 0.65 & 0.76 & 0.81 & 0.74 & 0.32 & 0.65 \\

\midrule
\multirow{2}{*}{Gemma-3-4B-IT}
& Vanilla
& 0.55 & 0.47 & 0.73 & 0.84 & 0.66 & 0.18 & 0.57 \\
& Steered
& 0.59 & 0.65 & 0.77 & 0.84 & 0.75 & 0.29 & 0.65 \\

\midrule
\multirow{2}{*}{Gemma-3-12B-PT}
& Vanilla
& 0.66 & 0.59 & 0.81 & 0.86 & 0.79 & 0.29 & 0.67 \\
& Steered
& 0.54 & 0.70 & 0.82 & 0.86 & 0.78 & 0.38 & 0.68 \\

\midrule
\multirow{2}{*}{Gemma-3-12B-IT}
& Vanilla
& 0.59 & 0.59 & 0.79 & 0.86 & 0.72 & 0.31 & 0.64 \\
& Steered
& 0.50 & 0.66 & 0.76 & 0.74 & 0.60 & 0.37 & 0.61 \\

\midrule
\multirow{2}{*}{Llama-3.1-8B}
& Vanilla
& 0.63 & 0.53 & 0.78 & 0.85 & 0.74 & 0.23 & 0.62 \\
& Steered
& 0.60 & 0.68 & 0.78 & 0.83 & 0.74 & 0.37 & 0.67 \\

\bottomrule
\end{tabular}
\caption{Emotion recognition performance averaged across three
additional semantically equivalent prompt templates. Vanilla denotes
the unsteered model.}
\label{tab:prompt_robustness}
\end{table*}

\begin{table*}[t]
\centering
\small
\setlength{\tabcolsep}{6pt}
\renewcommand{\arraystretch}{0.85}
\begin{tabular}{llccccc}
\toprule
\textbf{Model} & \textbf{Setting}
& \textbf{MMLU} & \textbf{ARC-C} & \textbf{BoolQ}
& \textbf{PIQA} & \textbf{Average} \\
\midrule

\multirow{2}{*}{Gemma-2-2B}
& Vanilla & 0.49 & 0.56 & 0.62 & 0.58 & 0.56 \\
& Steered & 0.49 & 0.56 & 0.62 & 0.58 & 0.56 \\

\midrule
\multirow{2}{*}{Gemma-2-9B}
& Vanilla & 0.68 & 0.87 & 0.86 & 0.84 & 0.81 \\
& Steered & 0.68 & 0.87 & 0.86 & 0.84 & 0.81 \\

\midrule
\multirow{2}{*}{Gemma-3-4B-PT}
& Vanilla & 0.56 & 0.72 & 0.76 & 0.73 & 0.69 \\
& Steered & 0.55 & 0.72 & 0.76 & 0.73 & 0.69 \\

\midrule
\multirow{2}{*}{Gemma-3-4B-IT}
& Vanilla & 0.58 & 0.77 & 0.81 & 0.79 & 0.74 \\
& Steered & 0.58 & 0.77 & 0.81 & 0.79 & 0.74 \\

\midrule
\multirow{2}{*}{Gemma-3-12B-PT}
& Vanilla & 0.71 & 0.88 & 0.86 & 0.83 & 0.82 \\
& Steered & 0.71 & 0.88 & 0.86 & 0.83 & 0.82 \\

\midrule
\multirow{2}{*}{Gemma-3-12B-IT}
& Vanilla & 0.71 & 0.90 & 0.86 & 0.87 & 0.84 \\
& Steered & 0.71 & 0.90 & 0.86 & 0.87 & 0.84 \\

\midrule
\multirow{2}{*}{Llama-3.1-8B}
& Vanilla & 0.63 & 0.79 & 0.75 & 0.78 & 0.74 \\
& Steered & 0.63 & 0.79 & 0.76 & 0.78 & 0.74 \\

\midrule
\multirow{2}{*}{\textbf{Average}}
& Vanilla
& \textbf{0.62} & \textbf{0.78} & \textbf{0.79}
& \textbf{0.77} & \textbf{0.74} \\
& Steered
& \textbf{0.62} & \textbf{0.78} & \textbf{0.79}
& \textbf{0.77} & \textbf{0.74} \\

\bottomrule
\end{tabular}
\caption{General-capability performance of vanilla and steered models.
Scores remain nearly unchanged after steering. The final rows report
averages across the seven models.}
\label{tab:general_capability_preservation}
\end{table*}

\subsection{Statistical Significance Tests}
\label{sec:statistic}

We conduct paired statistical significance tests across the seven evaluated models to compare sparse feature steering, few-shot prompting, and the zero-shot baseline. Because the comparisons involve paired measurements from the same models and test directional improvement hypotheses, we use exact one-sided Wilcoxon signed-rank tests. We apply Holm's correction separately within each block of five comparisons and report both unadjusted and Holm-adjusted \(p\)-values. We additionally report standardized paired effect sizes (\(d_z\)) based on the paired differences.

Table~\ref{tab:stat_tests} summarizes the results. Relative to the zero-shot baseline, sparse feature steering significantly improves macro-F1 on the main dataset from \(0.603\) to \(0.699\) (+15.9\%; Holm-adjusted \(p=0.0391\)). The improvements transfer to EmoEvent (\(0.410\) to \(0.463\), +12.9\%; Holm-adjusted \(p=0.0391\)) and XED (\(0.431\) to \(0.486\), +12.6\%; Holm-adjusted \(p=0.0391\)). Improvements are particularly strong for the weakest emotions identified in our earlier analyses. On the main dataset, \emph{disgust} F1 increases from \(0.416\) to \(0.683\) (+64.3\%; Holm-adjusted \(p=0.0391\)), while \emph{surprise} F1 increases from \(0.324\) to \(0.466\) (+43.6\%; Holm-adjusted \(p=0.0391\)).

Few-shot prompting also improves main-dataset macro-F1 from \(0.603\) to \(0.656\) (+8.8\%; Holm-adjusted \(p=0.0391\)) and \emph{disgust} F1 from \(0.416\) to \(0.551\) (+32.6\%; Holm-adjusted \(p=0.0391\)). Its improvement on \emph{surprise} is smaller and not statistically significant after correction (\(0.324\) to \(0.361\), +11.5\%; Holm-adjusted \(p=0.4453\)). Few-shot prompting also produces no significant improvement on EmoEvent or XED.

Compared with few-shot prompting, sparse feature steering significantly improves macro-F1 on the main dataset (\(0.656\) to \(0.699\), +6.5\%), EmoEvent (\(0.407\) to \(0.463\), +13.7\%), and XED (\(0.430\) to \(0.486\), +13.0\%). It also improves \emph{disgust} F1 by 23.8\% and \emph{surprise} F1 by 28.9\%. All five steering-versus-few-shot comparisons remain significant after Holm correction (Holm-adjusted \(p=0.0391\)).

\begin{table*}[t]
\centering
\small
\setlength{\tabcolsep}{5pt}
\begin{tabular}{llcccccc}
\toprule
Comparison & Metric & Mean A & Mean B & Rel. Change & $p$ & Holm-$p$ & $d_z$ \\
\midrule
\multirow{5}{*}{Steering vs. Baseline}
& Main Macro-F1
& 0.603 & 0.699 & +15.9\%
& 0.0078 & 0.0391 & 3.33 \\
& EmoEvent Macro-F1
& 0.410 & 0.463 & +12.9\%
& 0.0078 & 0.0391 & 1.71 \\
& XED Macro-F1
& 0.431 & 0.486 & +12.6\%
& 0.0156 & 0.0391 & 1.25 \\
& Disgust F1
& 0.416 & 0.683 & +64.3\%
& 0.0078 & 0.0391 & 2.99 \\
& Surprise F1
& 0.324 & 0.466 & +43.6\%
& 0.0078 & 0.0391 & 2.30 \\
\midrule
\multirow{5}{*}{Few-shot vs. Baseline}
& Main Macro-F1
& 0.603 & 0.656 & +8.8\%
& 0.0078 & 0.0391 & 1.55 \\
& EmoEvent Macro-F1
& 0.410 & 0.407 & -0.7\%
& 0.6016 & 1.0000 & -0.07 \\
& XED Macro-F1
& 0.431 & 0.430 & -0.3\%
& 0.5859 & 1.0000 & -0.03 \\
& Disgust F1
& 0.416 & 0.551 & +32.6\%
& 0.0078 & 0.0391 & 1.17 \\
& Surprise F1
& 0.324 & 0.361 & +11.5\%
& 0.1484 & 0.4453 & 0.47 \\
\midrule
\multirow{5}{*}{Steering vs. Few-shot}
& Main Macro-F1
& 0.656 & 0.699 & +6.5\%
& 0.0078 & 0.0391 & 3.85 \\
& EmoEvent Macro-F1
& 0.407 & 0.463 & +13.7\%
& 0.0156 & 0.0391 & 1.12 \\
& XED Macro-F1
& 0.430 & 0.486 & +13.0\%
& 0.0312 & 0.0391 & 1.02 \\
& Disgust F1
& 0.551 & 0.683 & +23.8\%
& 0.0078 & 0.0391 & 2.84 \\
& Surprise F1
& 0.361 & 0.466 & +28.9\%
& 0.0078 & 0.0391 & 1.39 \\
\bottomrule
\end{tabular}
\caption{
Paired statistical significance tests across seven models comparing sparse feature steering, few-shot prompting, and the zero-shot baseline. We report exact one-sided Wilcoxon signed-rank tests, unadjusted \(p\)-values, Holm-adjusted \(p\)-values, and standardized paired effect sizes (\(d_z\)). Holm's correction is applied separately within each block of five comparisons. Sparse feature steering yields significant and transferable improvements, particularly for weakly represented emotions such as \emph{disgust} and \emph{surprise}.
}
\label{tab:stat_tests}
\end{table*}

\begin{table*}[t]
\centering
\scriptsize
\begin{tabular}{p{2.0cm} c p{5.1cm} p{5.1cm}}
\toprule
\textbf{Model} & \textbf{\#} & \textbf{Cluster 0} & \textbf{Cluster 1} \\
\midrule

\multirow{8}{*}{\makecell[l]{Gemma-2-2B \\ (joy)}}
& 1 & Winning a race after many weeks of training. & Finally getting a first 5-star reward before the set ends. \\
& 2 & Realizing one's desires and aims. & Obtaining money needed for a planned purchase. \\
& 3 & Going to the aquarium with a boyfriend and enjoying the animals together. & Being accepted as supervisor for a student teacher. \\
& 4 & Finally feeling well again after a long battle with depression. & Opening the TE score envelope and getting into Physiotherapy. \\
& 5 & Singing with a band in front of a large school audience. & Achieving a first 6 at university after lower previous scores. \\
& 6 & Finishing a long hike with a sense of accomplishment. & Receiving a positive examination result for a Psychology degree. \\
& 7 & Winning in a video game. & Being accepted into a first-choice university. \\
& 8 & Favorite national team winning an international sports event. & Being accepted to Villanova University. \\

\midrule

\multirow{8}{*}{\makecell[l]{Llama-3.1-8B \\ (joy)}}
& 1 & Recovering from depression and finally feeling well again. & Receiving a university acceptance letter after waiting nearly a year for exam results. \\
& 2 & Reuniting with a long-distance partner after COVID border restrictions ended. & Winning a close tennis match after coming from behind. \\
& 3 & Jumping into a pond during a summer camp night swim with a friend. & Passing school leaving exams and being selected for college. \\
& 4 & Beating Virtual Clash for the first time after two years. & Watching one's team win the cup. \\
& 5 & Singing and laughing during a sunset drive in South Africa with loved ones. & Achieving a first 6 at university after a long series of lower scores. \\
& 6 & Winning an auction for a long-desired item. & Being accepted into a first-choice university. \\
& 7 & Recovering from bulimia and depression and finally feeling like dancing again. & Watching a favorite football team win the cup final. \\
& 8 & Traveling to visit one's grandmother and enjoying the seaside with family. & Watching one's team win. \\

\midrule

\multirow{8}{*}{\makecell[l]{Gemma-2-2B \\ (disgust)}}
& 1 & Seeing someone pick their nose and eat it. & Detesting the cruel behavior of a fellow soldier. \\
& 2 & Watching people eat with bad manners such as smacking or belching. & Feeling revulsion toward someone behaving like a scoundrel. \\
& 3 & Accidentally eating a spoiled pistachio nut. & Being approached and insulted by a drunk man in the underground station. \\
& 4 & Eating unpleasant food after a cold and exhausting mountain climb. & Reading a book with obscene and tasteless content. \\
& 5 & Seeing someone eat live worms. & Feeling disgust toward brutal violence and physical aggression. \\
& 6 & Watching someone noisily gulp food in a cafeteria. & Speaking to a man avoiding responsibility for his family's suffering. \\
& 7 & Eating a spoiled brownie from the fridge. & Seeing someone spit on the floor in the street. \\
& 8 & Seeing someone use meat after dropping it on the floor. & Seeing someone spit inside a shopping centre. \\

\bottomrule
\end{tabular}
\caption{Top activating examples for joy and disgust sparse feature clusters across models. Joy clusters consistently separate experiential/social joy from achievement- and reward-based joy, while disgust clusters separate physical contamination from social or norm-based aversion.}
\label{tab:emotion_cluster_examples}
\end{table*}

\begin{table*}[t]
\centering
\small
\setlength{\tabcolsep}{3pt}
\begin{tabular}{llrrrrr}
\toprule
\textbf{Model} & \textbf{Emotion} & \textbf{X$\to$M act. drop} & \textbf{X$\to$Y loss} & \textbf{M$\to$Y loss} & \textbf{Restore $\Delta$logit} & \textbf{Recovery} \\
\midrule

\multirow{6}{*}{google/gemma-2-2b}
& anger & 21.08 & 0.62 & 0.81 & 0.55 & 0.85 \\
& disgust & 5.12 & 0.42 & 0.61 & 0.29 & 0.68 \\
& fear & 25.60 & 1.28 & 1.68 & 1.10 & 0.87 \\
& joy & 10.71 & 0.71 & 0.90 & 0.62 & 0.80 \\
& sadness & 10.06 & 0.70 & 0.91 & 0.60 & 0.79 \\
& surprise & 14.84 & 1.11 & 1.39 & 0.87 & 0.84 \\
\midrule

\multirow{6}{*}{google/gemma-2-9b}
& anger & 13.25 & 0.45 & 0.57 & 0.40 & 0.87 \\
& disgust & 6.12 & 0.57 & 0.66 & 0.45 & 0.72 \\
& fear & 26.10 & 0.88 & 1.09 & 0.82 & 0.89 \\
& joy & 11.12 & 0.75 & 0.92 & 0.64 & 0.84 \\
& sadness & 8.66 & 0.71 & 0.93 & 0.64 & 0.81 \\
& surprise & 11.61 & 1.04 & 1.28 & 1.02 & 0.87 \\
\midrule

\multirow{6}{*}{google/gemma-3-12b-IT}
& anger & 2156.79 & 1.40 & 1.61 & 1.24 & 0.91 \\
& disgust & 1194.09 & 1.18 & 1.37 & 0.85 & 0.75 \\
& fear & 2922.09 & 1.74 & 1.95 & 1.50 & 0.95 \\
& joy & 1779.15 & 1.74 & 2.14 & 1.39 & 0.81 \\
& sadness & 2805.30 & 1.53 & 1.89 & 1.18 & 0.88 \\
& surprise & 2102.53 & 1.98 & 2.10 & 1.56 & 0.82 \\
\midrule

\multirow{6}{*}{google/gemma-3-12b-PT}
& anger & 1676.88 & 0.61 & 0.70 & 0.46 & 0.84 \\
& disgust & 833.75 & 0.47 & 0.52 & 0.30 & 0.72 \\
& fear & 2643.63 & 0.99 & 1.15 & 0.98 & 1.05 \\
& joy & 1734.65 & 1.07 & 1.19 & 0.90 & 0.91 \\
& sadness & 1747.14 & 0.79 & 1.08 & 0.61 & 0.78 \\
& surprise & 1958.14 & 1.24 & 1.36 & 1.06 & 0.83 \\
\midrule

\multirow{6}{*}{google/gemma-3-4b-IT}
& anger & 590.45 & 2.06 & 2.38 & 1.62 & 0.68 \\
& disgust & 258.13 & 2.10 & 2.44 & 1.31 & 0.51 \\
& fear & 678.87 & 2.36 & 2.77 & 1.74 & 0.73 \\
& joy & 526.19 & 2.82 & 3.38 & 2.04 & 0.73 \\
& sadness & 439.46 & 1.93 & 2.30 & 1.31 & 0.59 \\
& surprise & 597.35 & 2.70 & 3.00 & 1.94 & 0.73 \\
\midrule

\multirow{6}{*}{google/gemma-3-4b-PT}
& anger & 330.26 & 0.69 & 0.81 & 0.53 & 0.73 \\
& disgust & 238.28 & 1.14 & 1.34 & 0.73 & 0.61 \\
& fear & 848.02 & 1.39 & 1.64 & 1.22 & 0.90 \\
& joy & 403.29 & 1.03 & 1.20 & 0.70 & 0.73 \\
& sadness & 390.71 & 1.04 & 1.27 & 0.80 & 0.73 \\
& surprise & 276.03 & 0.85 & 0.97 & 0.63 & 0.59 \\
\midrule

\multirow{6}{*}{meta-llama/Llama-3.1-8B}
& anger & 1.91 & 0.72 & 0.91 & 0.60 & 0.72 \\
& disgust & 1.05 & 0.52 & 0.63 & 0.32 & 0.54 \\
& fear & 3.35 & 1.50 & 1.82 & 1.47 & 0.99 \\
& joy & 1.74 & 0.93 & 1.09 & 0.74 & 0.79 \\
& sadness & 1.53 & 0.93 & 1.16 & 0.77 & 0.75 \\
& surprise & 1.06 & 0.74 & 0.90 & 0.68 & 0.73 \\
\midrule

\multirow{6}{*}{ALL\_MODELS\_AVG}
& anger & 684.37 & 0.93 & 1.11 & 0.77 & 0.80 \\
& disgust & 362.36 & 0.91 & 1.08 & 0.61 & 0.65 \\
& fear & 1021.09 & 1.45 & 1.73 & 1.26 & 0.91 \\
& joy & 638.12 & 1.29 & 1.54 & 1.01 & 0.80 \\
& sadness & 771.84 & 1.09 & 1.36 & 0.85 & 0.76 \\
& surprise & 708.79 & 1.38 & 1.57 & 1.11 & 0.77 \\
\bottomrule
\end{tabular}
\caption{Mean adjacent-layer group path-patching results per emotion across models.}
\label{tab:all_models_adjacent_group_emotion_mean}
\end{table*}

\end{document}